\documentclass[journal]{IEEEtran}
\usepackage{easyReview}
\usepackage{comment}

\usepackage{multirow}
\usepackage{graphicx}
\usepackage{array} 
\usepackage{float}
\usepackage{subcaption}
\usepackage{epstopdf}
\usepackage{amsmath}
\usepackage{amsfonts}
\usepackage{algorithm}
\usepackage{algorithmic}
\usepackage{makecell}
\usepackage{booktabs}
\setcellgapes{3pt}

\setcellgapes{3pt}

\usepackage[justification=centering]{caption}
\usepackage{color}
\usepackage{multirow}
\usepackage{tabularx}
\usepackage{adjustbox}  
\usepackage{amssymb}

\makeatletter

\newcommand{\Rmnum}[1]{\expandafter\@slowromancap\romannumeral #1@}
\makeatother

\usepackage{cite}

\ifCLASSINFOpdf
\else
\fi
\begin{document}
\bstctlcite{IEEEexample:BSTcontrol}
%
\title{TokenCom: Vision-Language Model for Multimodal and Multitask Token Communications}

\author{Feibo Jiang, \textit{Senior Member, IEEE}, Siwei Tu, Li Dong, Xiaolong Li, Kezhi Wang, \textit{Senior Member, IEEE}, Cunhua Pan,  \textit{Senior Member, IEEE}, Zhu Han, \textit{Fellow, IEEE}, Jiangzhou Wang, \textit{Fellow, IEEE}
\thanks{This work was supported in part by the National Natural Science Foundation
	of China under Grant 62572184; in part by Hunan Provincial Natural Science
	Foundation of China under Grant 2024JJ5270 and Grant 2025JJ50365; in part by Changsha
	Natural Science Foundation under Grant kq2402098 and Grant kq2402162.

		Feibo Jiang (jiangfb@hunnu.edu.cn) is with Hunan Provincial Key Laboratory of Intelligent Computing and Language Information Processing, Hunan Normal University, Changsha, China.
		
		Siwei Tu (tusiwei@hunnu.edu.cn) is with School of Information Science and Engineering, Hunan Normal University, Changsha, China.
		
		Li Dong (Dlj2017@hunnu.edu.cn) is with Changsha Social Laboratory of Artificial Intelligence, Hunan University of Technology and Business, Changsha, China.
		
		Xiaolong Li (lxl@hutb.edu.cn) is with the School of Computer Science, Hunan University of Technology and Business, Changsha, China.
		
		Kezhi Wang (Kezhi.Wang@brunel.ac.uk) is with the Department of Computer Science, Brunel University London, UK.
		
		
		Cunhua Pan (cpan@seu.edu.cn) is with the National Mobile Communications Research Laboratory, Southeast University, Nanjing 210096, China.
		
		Zhu Han (zhan2@uh.edu) is with the Department of Electrical and Computer Engineering, University of Houston, Houston, TX, USA.
		
		Jiangzhou Wang (j.z.wang@seu.edu.cn) is with the National Mobile Communications Research Laboratory, Southeast University, Nanjing, China, and also with the Purple Mountain Laboratories, Nanjing, China.
		
	}
}
\markboth{Submitted for Review}%
{Shell \MakeLowercase{\textit{et al.}}: Bare Demo of IEEEtran.cls for IEEE Journals}
%



\maketitle


\begin{abstract}

Visual–Language Models (VLMs), with their strong capabilities in image and text understanding, offer a solid foundation for intelligent communications. However, their effectiveness is constrained by limited token granularity, overlong visual token sequences, and inadequate cross-modal alignment. To overcome these challenges, we propose TaiChi, a novel VLM framework designed for token communications. TaiChi adopts a dual-visual tokenizer architecture that processes both high- and low-resolution images to collaboratively capture pixel-level details and global conceptual features. A Bilateral Attention Network (BAN) is introduced to intelligently fuse multi-scale visual tokens, thereby enhancing visual understanding and producing compact visual tokens. In addition, a Kolmogorov–Arnold Network (KAN)-based modality projector with learnable activation functions is employed to achieve precise nonlinear alignment from visual features to the text semantic space, thus minimizing information loss. Finally, TaiChi is integrated into a multimodal and multitask token communication system equipped with a joint VLM–channel coding scheme. Experimental results validate the superior performance of TaiChi, as well as the feasibility and effectiveness of the TaiChi-driven token communication system.
\end{abstract}


\IEEEpeerreviewmaketitle

\section{Introduction}

\IEEEPARstart{I}{n} recent years, the communication paradigm has undergone a profound transformation, shifting from bit-level accurate replication toward the efficient transmission of tokens, a paradigm referred to as token communication. The core objective of this paradigm is to enable the receiver to accurately reconstruct and interpret the sender’s intent based on the transmitted tokens, a requirement that becomes increasingly critical with the rapid growth of multimodal data such as text and images\cite{10670195}. To address this challenge, Visual-Language Models (VLMs) have emerged as a key enabling technology, as they are capable of extracting and generating high-level token representations from multimodal data and therefore exhibit great potential to serve as a cornerstone of next-generation intelligent communication systems. By integrating powerful visual models with Large Language Models (LLMs), VLMs can map visual signals into discrete or continuous token representations that are aligned with the semantic space of LLMs, thereby laying the foundation for an advanced multimodal tokenizer and alignment\cite{liang2025vision}. However, directly applying existing VLMs to token communication still faces several challenges as follows.

\subsubsection{Limited Token Granularity}
In token communication, the primary task of a tokenizer is to comprehensively capture the full ``meaning" of the source visual information. Most current VLMs rely on a single, low-resolution visual tokenizer, which is unable to simultaneously capture both global and detailed information. This limitation prevents the visual tokenizer from adaptively selecting whether to transmit global or detailed features based on the specific requirements of communication tasks \cite{zhao2025cobra}.

\subsubsection{Overlong Visual Token Sequences}
To enable LLMs to process visual information, high-dimensional visual features are typically encoded into long sequences of visual tokens. However, excessively long token sequences are difficult to fully model under limited context length and computational resource constraints, which tends to dilute or overlook critical information, thereby compromising the integrity of multimodal feature representations. Moreover, long token sequences incur substantial communication overhead \cite{li2022blip}, forcing models to truncate or compress visual tokens, which further exacerbates token information loss during transmission.

\subsubsection{Inadequate Cross-Modal Alignment}
Projectors align visual tokens with the text semantic space, typically using Multi-Layer Perceptrons (MLPs) for semantic alignment. However, the fixed activation functions of MLPs make it difficult to efficiently model the complex nonlinear relationships between the heterogeneous modalities of vision and language. Moreover, the ``spectral bias" problem inherent in MLPs, which prioritizes the learning of low-frequency information in visual tokens, may lead to the omission of high-frequency visual information during the projection process \cite{yang2024kolmogorov}.

To address the aforementioned challenges, we propose TaiChi, a high-quality VLM framework that can be applied to token communications, which makes the following key contributions:

\setcounter{subsubsection}{0}

\subsubsection{Dual-Visual Tokenizer}
To comprehensively capture image features, we design a dual-visual tokenizer architecture. The architecture employs a Vision Transformer (ViT)-based tokenizer to extract global  features from low-resolution images, while incorporating a convolution-based tokenizer to capture fine-grained local details from high-resolution images. This complementary design establishes a rich foundation of visual representations, enabling the system to flexibly encode and transmit image information at varying levels of granularity according to task requirements.

\subsubsection{Bilateral Attention Network}
\textcolor{black}{To achieve efficient and bidirectional fusion between high-level and low-level image features, we propose a Bilateral Attention Network (BAN). Specifically, BAN employs two complementary attention branches: one branch uses high-level features as queries to extract fine-grained details from low-level representations; while the other branch uses low-level features as queries to capture global conceptual information from high-level representations. Through BAN, the network enables mutual enhancement between global features and local details. Afterward, a pooling and aggregation operation aligns the spatial resolutions of both streams, and their outputs are fused to generate refined visual representations. The enhanced visual tokens jointly capture global features and salient local details, while substantially reducing the total number of visual tokens.}

\subsubsection{KAN-based Projector}
To achieve efficient alignment between the visual and language modalities, we introduce a Kolmogorov-Arnold Network (KAN) as the modality projector. Unlike MLPs, KAN incorporates learnable activation functions within the network architecture, enabling automatic learning of the optimal nonlinear mapping between features based on data. Its superior function approximation ability and learning efficiency allow for smoother and more accurate mapping of visual tokens at different scales to the semantic space of LLMs, effectively reducing information loss during modality alignment and ensuring high fidelity.

\subsubsection{TaiChi-Driven Token Communication}
We innovatively design an advanced multimodal and multitask token communication system with TaiChi as the core component. We adopt a multimodal and multitask instruction tuning strategy, which regularizes different communication tasks into a unified instruction template using natural language commands. This enables the VLM to efficiently learn and understand multiple multimodal tasks. To ensure robust transmission of token information over physical channels, we further propose a joint VLM-channel coding scheme. This scheme optimizes both feature extraction and channel coding processes, allowing the system to maintain outstanding performance even in the presence of channel noise and interference, significantly enhancing the overall robustness.

The remainder of this paper is organized as follows: Section \uppercase\expandafter{\romannumeral2} reviews related work. Section \uppercase\expandafter{\romannumeral3} introduces the TaiChi framework. Section \uppercase\expandafter{\romannumeral4} presents the token communication system based on TaiChi. Section \uppercase\expandafter{\romannumeral5} describes the training methods and data. Section \uppercase\expandafter{\romannumeral6} provides experimental setups and results, and finally Section \uppercase\expandafter{\romannumeral7} concludes the paper.

\section{Related Work}
\subsection{Vision Language Model}
The LLaVA model proposed by Liu et al. \cite{liu2023visual} established the foundational framework for modern VLM through its concise architecture, which connects the visual encoder with an LLM. Liu also introduced the pioneering ``feature alignment pretraining + instruction fine-tuning" two-stage training method. Zhu et al.'s MiniGPT-4 \cite{zhu2023minigpt} revealed that relying solely on large-scale noisy data for pretraining leads to poor quality of generated content. They demonstrated the critical role of data quality in improving model reliability by incorporating an additional fine-tuning stage using a high-quality, small-scale dataset. To address the substantial computational costs associated with processing high-resolution images, Bai et al. \cite{bai2025qwen2} proposed the Qwen-VL series, which introduced a cross-attention-based visual adapter. This effectively compresses long visual feature sequences into fixed-length short sequences, significantly improving computational efficiency. Additionally, Chen et al.'s InternVL series \cite{chen2024internvl} fundamentally revolutionized the training paradigm by introducing the concept of unified pretraining, enabling the model to jointly learn language and multimodal capabilities from the outset. 

\subsection{VLM-based Semantic Communication}

Ni et al. \cite{ni2025multi} proposed a Multi-Task SemCom (MTSC) architecture based on large models. The architecture also improves the accuracy of semantic extraction and content generation by integrating retrieval-augmented generation schemes. Du et al. \cite{10976624} studied a vehicle AI assistant based on Multimodal Large Language Models (MLLM) and proposed a task-oriented semantic communication framework, significantly improving question-answering accuracy under poor Signal-to-Noise Ratio (SNR) conditions. Zhang et al. \cite{zhang2025multimodal} introduced a novel MLLM-integrated semantic communication framework, called MLLM-SC. This framework optimizes adaptive bandwidth allocation and high-quality content reconstruction or generation through an importance-aware semantic encoder and a resource-adaptive semantic decoder. Cao et al. \cite{10615340} proposed a Privacy-preserving Semantic Communication scheme on MLLM (MLLM-PSC), which utilizes the semantic understanding capabilities of MLLMs to achieve a unified conversion of multimodal data into text semantics. 

\subsection{Token Communication}
Qiao et al. \cite{11175596} proposed Token Communications (TokCom), a large-model-driven paradigm for cross-modal semantic communications. By representing multimodal data as unified discrete tokens and leveraging pre-trained MLLMs for masked/next-token prediction, TokCom reconstructs missing tokens at extremely low bitrates, establishing “tokens as semantics, semantics as communication”.
Liu et al. \cite{11149073} designed Text-Guided TokCom, the first system to employ textual tokens as side information for wireless image transmission. 
Zhang et al. \cite{zhang2025tokcom} presented TokCom-UEP, revealing that equal-error protection wastes redundancy because 1-D token sequences exhibit inherent semantic hierarchy. Using Expanding-Window Fountain codes, they match protection level to token importance, confirming that TokCom should be priced by semantic significance rather than by raw bits.
Qiao et al. \cite{qiao2025token} introduced Token-Domain Multiple Access (ToDMA), the first multi-access framework that formalizes TokCom.

In summary, although VLMs and their applications in semantic communication have achieved notable progress, they still suffer from inherent limitations in token-level visual representation and subsequent alignment. Moreover, due to the large parameter scale of VLMs, end-to-end design and training for token communication have not yet been adequately explored. To address these issues, this study proposes TaiChi-driven token communication as an effective solution.

\section{TaiChi Framework}
\begin{figure*}[htpb]
	\centering
	\includegraphics[width=17cm]{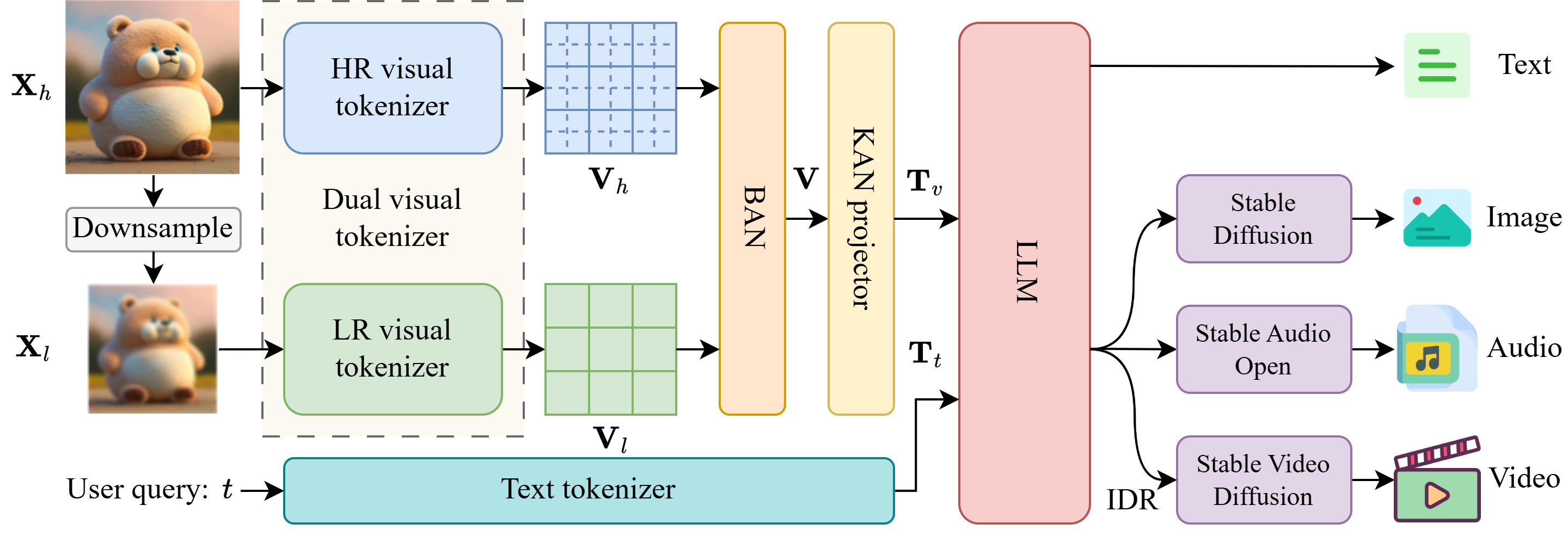}
	\caption{The structure of the proposed TaiChi framework.}
	\label{fig:main}
\end{figure*}

We first introduce the TaiChi framework, as illustrated in Fig. \ref{fig:main}, which comprises a dual-visual tokenizer, a BAN, a KAN-based projector, an LLM, and a diffusion model. The two visual tokenizers separately process high-resolution and low-resolution images, performing hierarchical encoding while balancing global features and fine-grained details. Subsequently, the BAN integrates micro-level details into the global features under receptive fields of varying scales, effectively merging the two sets of visual features into a compact and efficient set of visual tokens. These visual tokens are then projected into the LLM’s text semantic space through the KAN-based projector and, together with the original text tokens, are fed into the LLM to generate textual outputs.
When multimodal outputs such as image, audio, or video are required, TaiChi further employs a method called Instructional Diffusion Refinement (IDR) to guide the diffusion models in producing high-quality multimodal content.

\subsection{Dual-Visual Tokenizers}
To address the limitations of a single visual tokenizer in representing image information, specifically the inherent trade-off between capturing high-level features and preserving fine-grained details, the TaiChi framework designs an innovative dual-visual tokenizer architecture. This architecture aims to decouple the global conceptual features of images from the perception of local details, generating complementary visual features through two parallel processing paths. As shown in Fig. \ref{fig:main}, the original High Resolution (HR) image $\textbf{X}_h$ is input into an HR visual tokenizer optimized for capturing details, while the downsampled Low Resolution (LR) version $\textbf{X}_l$ is sent to an LR visual tokenizer focused on global features. This design allows for the separate extraction of the detail feature stream $\textbf{V}_h$ and the global feature stream $\textbf{V}_l$, which together form a comprehensive and information-rich foundation for visual representation.

\subsubsection{LR visual tokenizer}
To obtain robust global features of the input image, we select the ViT model \cite{dosovitskiy2020image} as our LR visual tokenizer. ViT directly applies the Transformer architecture to image tasks, using the self-attention mechanism to capture long-range dependencies between various regions of the image. This allows ViT to extract rich visual features that enhance the model’s ability to comprehend the global features of the image. In the TaiChi framework, we leverage this capacity to effectively extract the core themes and conceptual information from low-resolution images, thereby generating a powerful global visual feature stream.

As shown in Fig. \ref{fig:ve}, the ViT model consists of a stack of $L$ layers of Transformer encoders, where the multi-head attention mechanism excels at capturing global dependencies between various positions in the image sequence, while the MLP independently processes the representations of each position to capture fine-grained local features. Additionally, residual connections help the information bypass layers, alleviating issues such as vanishing gradients and network degradation, while layer normalization standardizes the output of each sublayer, improving training stability and convergence speed \cite{vaswani2017attention}. Specifically, the outputs of each layer’s multi-head attention and the MLP can be represented as:
\begin{equation}
    \textbf{M}_{mha, i} =
\begin{cases}
\text{MHA}(\text{LN}(\textbf{X}_{l})) + \textbf{X}_{l}, & i=1, \\
\text{MHA}(\text{LN}(\textbf{M}_{ff, i-1})) + \textbf{M}_{ff, i-1}, & 2 \le i \le L,
\end{cases}
\end{equation}
\begin{equation}
\begin{split}
\textbf{M}_{ff, i} = \text{GeLU}(\textbf{W}_{b, f} \cdot \text{LN}(\textbf{M}_{mha, i}) + \textbf{b}_{b, f}) + \textbf{M}_{mha, i},\\ \quad 1 \le i \le L,
\end{split}
\end{equation}
where $\text{MHA}(\cdot)$ represents the multi-head attention operator, as shown in Eq. (\ref{eq:MHA}), and $\text{LN}(\cdot)$ refers to the layer normalization operator. $\textbf{W}_{b, f}$ and $\textbf{b}_{b, f}$ are the weights and biases in the MLP, while $\text{GeLU}(\cdot)$ denotes the activation function in the MLP.
\begin{equation}
\text{MHA}(\textbf{X}) = \text{Softmax}\left(\frac{\textbf{W}_Q \textbf{X} (\textbf{W}_K \textbf{X})^T}{\sqrt{d_k}}\right)\textbf{W}_V \textbf{X},
\label{eq:MHA}
\end{equation}
where $\text{Softmax}(\cdot)$ is the normalization operator, while $\textbf{W}_Q, \textbf{W}_K$ and $\textbf{W}_V$ are the matrices that map the input to the query, key, and value, respectively. $\sqrt{d_k}$ represents the feature dimension of the input, and it is used to prevent gradient vanishing\cite{dosovitskiy2020image}.

\subsubsection{HR visual tokenizer}
To overcome the limitations of LR visual tokenizers in capturing fine-grained features, we introduce modern Convolutional Neural Networks (CNNs) as HR visual tokenizers \cite{liu2022convnet}. CNNs, through their inherent convolution operations, are highly effective in extracting local detail information from images, especially from HR images, capturing textures, edges, and other high-frequency features. The convolution operations are adept at capturing local patterns at various scales within the image and leveraging local dependencies between pixels to generate rich visual features. By applying convolutional operations to HR images $\textbf{X}_h$, we are able to fully exploit the multi-scale local information extracted by the CNN, thereby enhancing the model's ability to perceive fine details.

\begin{figure}[htpb]
	\centering
	\includegraphics[width=7cm]{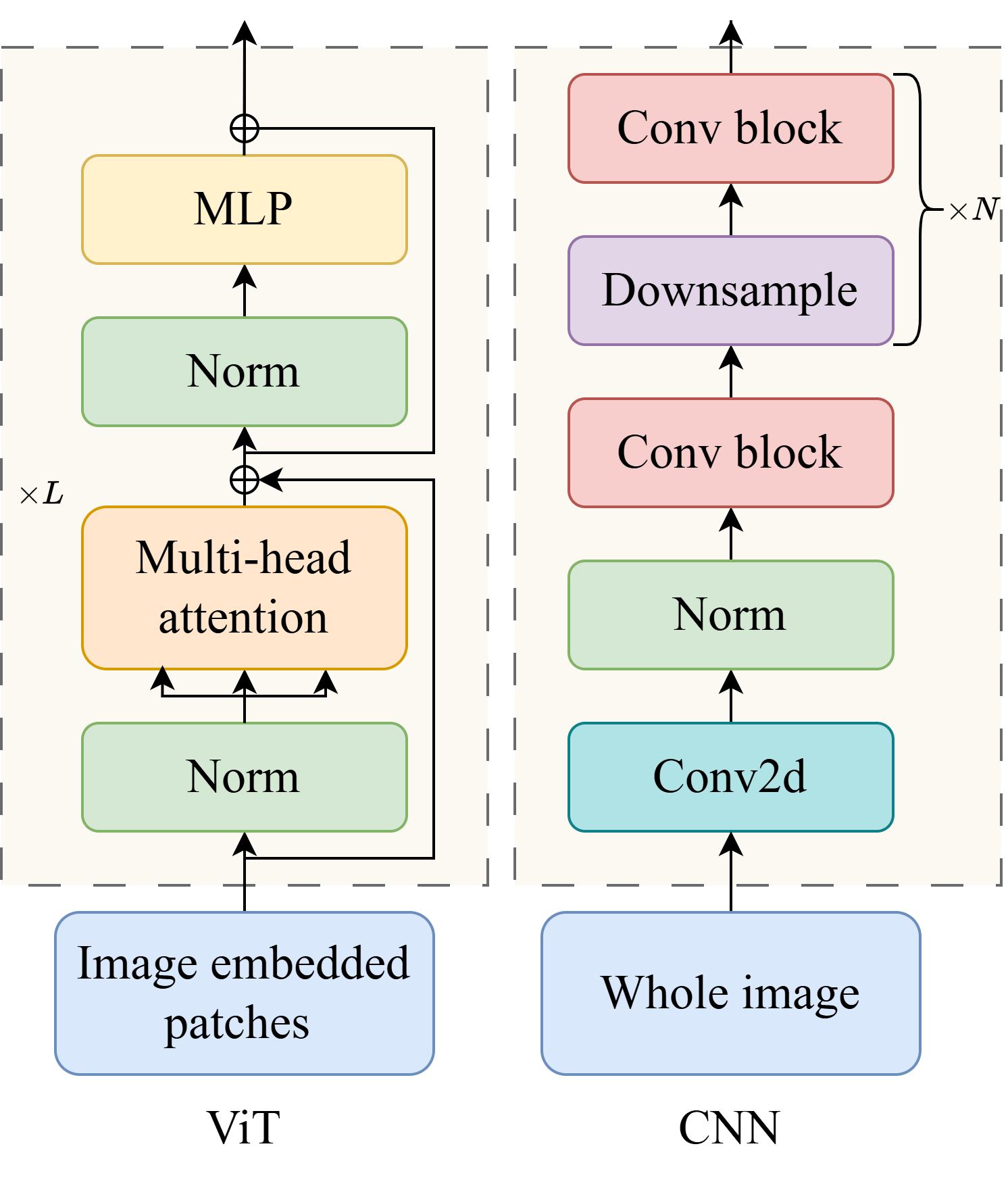}
	\caption{Dual-visual tokenizer architecture.}
	\label{fig:ve}
\end{figure}

As shown in Fig. \ref{fig:ve}, the CNN utilizes a series of standard modules, such as convolutional blocks, and downsampling operations. Convolutional blocks use filter sliding windows to extract local features from the image. Furthermore, CNNs typically include normalization layers to enhance the stability of training and mitigate issues such as vanishing gradients. Through the combination of these modules, the CNN gradually extracts and integrates multi-level features from the image across multiple processing stages, forming a robust detail feature stream $\textbf{V}_h$:
\begin{equation}
	\begin{split}
		\textbf{V}_h =\text{Conv}(\text{Down}(\text{Conv}(\text{LN}(\text{C}_{2d}(\textbf{X}_h))))),
	\end{split}
\end{equation}
where $\text{Conv}(\cdot)$ denotes convolutional blocks, $\text{Down}(\cdot)$ is downsample operation and $\text{C}_{2d}(\cdot)$ is two-dimensional convolution operation \cite{liu2022convnet}.

Ultimately, the LR visual tokenizer provides global feature encoding about ``what" the image represents, while the HR visual tokenizer supplements this with detailed feature encoding regarding ``how" it is composed and ``where" it is located. Together, these two visual tokenizers work synergistically, offering TaiChi a comprehensive visual understanding capability far beyond that of a single visual tokenizer.

\subsection{\textcolor{black}{Bilateral Attention Network}}

After the dual visual tokenizers separately extract the high-level and low-level features, the key challenge lies in how to efficiently and effectively integrate them. 
To address this, we propose the BAN, a bidirectional attention fusion module designed to enable fine-grained and semantically coherent interaction between the two feature streams. 
Specifically, BAN consists of two complementary attention branches: 
1) a \textit{top-down branch}, where high-level features $\textbf{V}_h$ act as queries to selectively extract relevant local details from the low-level feature map $\textbf{V}_l$; and 
2) a \textit{bottom-up branch}, where low-level detailed features $\textbf{V}_l$ serve as queries to capture global conceptual information from the high-level feature stream $\textbf{V}_h$. 
Through this bidirectional attention interaction, the model achieves feature refinement and detail recovery simultaneously.

Unlike conventional global Cross-Attention (CA), which computes dense interactions between all tokens, BAN performs localized spatial correspondence attention. 
Each token only attends to its spatially aligned region in the other feature stream, reducing redundant computation while preserving spatial consistency. 
This precise correspondence based fusion enables the model to enhance important regions such as object edges, boundaries, and fine textures, while effectively suppressing irrelevant background noise. As a result, both the discriminative ability and computational efficiency are improved.

\subsubsection{Top-down localized attention}
In the top-down branch, high-level features guide the extraction of spatially corresponding fine details. 
Let $\textbf{V}_h$ and $\textbf{V}_l$ denote the high- and low-level feature maps, respectively. 
For each spatial position $i$, the high-level token $\textbf{V}_{h,i}$ queries the spatially aligned detailed token $\textbf{V}_{l,i}$ to obtain an enhanced token representation $\textbf{V}_{h,i}^{\prime}$:
\begin{equation}
	\textbf{V}_{h,i}^{\prime} = 
	\text{Softmax}\!\left(
	\frac{
		(\textbf{W}_{Q,h} \textbf{V}_{h,i})(\textbf{W}_{K,h} \textbf{V}_{l,i})^{T}
	}{\sqrt{d_k}}
	\right) 
	\textbf{W}_{V,h} \textbf{V}_{l,i},
	\label{eq:ban_topdown}
\end{equation}
where $\textbf{W}_{Q,h}$, $\textbf{W}_{K,h}$, and $\textbf{W}_{V,h}$ are the query, key, and value matrices used for top-down localized attention, respectively.

This one-to-one correspondence enables each token concept to ``inspect'' its local region and integrate the most relevant fine-grained details.

\subsubsection{Bottom-up localized attention}
Conversely, the bottom-up branch allows low-level detailed features to extract semantic guidance from high-level representations. 
Similar to the aforementioned, at the fine-grained level, a single token $\textbf{V}_{l,i}$ interacts only with its aligned location, and the enhanced token representation $\textbf{V}_{l,i}^{\prime}$ can be expressed as:
\begin{equation}
	\textbf{V}_{l,i}^{\prime} =
	\text{Softmax}\!\left(
	\frac{
		(\textbf{W}_{Q,l} \textbf{V}_{l,i})(\textbf{W}_{K,l} \textbf{V}_{h,i})^{T}
	}{\sqrt{d_k}}
	\right) 
	\textbf{W}_{V,l} \textbf{V}_{h,i},
	\label{eq:ban_bottomup}
\end{equation}
where $\textbf{W}_{Q,l}$, $\textbf{W}_{K,l}$, and $\textbf{W}_{V,l}$ are the query, key, and value matrices used for bottom-up localized attention, respectively.

\subsubsection{Feature fusion}
\begin{figure}[htpb]
	\centering
	\includegraphics[width=9cm]{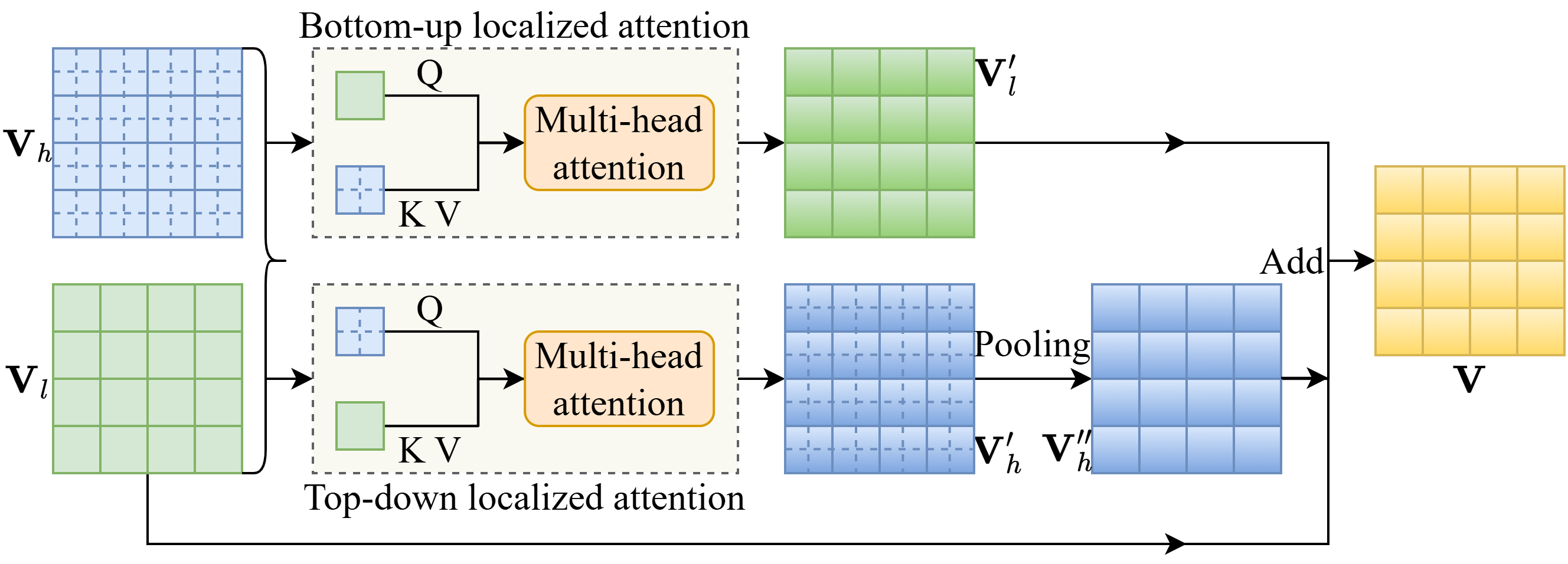}
	\caption{The computation pipeline of BAN.}
	\label{fig:ban}
\end{figure}

After performing bidirectional attention, BAN fuses the enhanced high and low-level features through pooling and residual aggregation, as illustrated in Fig.~\ref{fig:ban}. 
The final fused representation $\textbf{V}$ is computed as:
\begin{equation}
	\textbf{V} = \textbf{V}_l
	+ \textbf{V}_l^{\prime} 
	+ \text{Pooling}(\textbf{V}_h^{\prime}),
	\label{eq:ban_fusion}
\end{equation}
where $\textbf{V}_l^{\prime}$ and $\textbf{V}_h^{\prime}$ denote the outputs of the two attention branches, and $\text{Pooling}(\cdot)$ aligns spatial resolutions before residual fusion. 
This process produces a set of high-quality visual tokens that jointly encode semantic abstraction and detailed fidelity, providing compact yet expressive inputs for subsequent cross-modal alignment.

\subsection{KAN-based Projector}
In the process of integrating visual information into LLMs, the modality projector plays a crucial role. Its core task is to map the high-dimensional, complex visual features extracted by the visual tokenizer into a word embedding space that the LLM can understand. Traditionally, MLPs have been widely adopted due to their simple structure and theoretical properties as universal function approximators. However, MLPs have practical limitations in terms of expressive power, particularly when using piecewise linear activation functions (such as ReLU), which struggle to efficiently approximate functions with complex periodic or high-frequency characteristics. Additionally, the gradient descent-based training method leads to the ``spectral bias" problem, where the network tends to prioritize learning the low-frequency components of the target function, with convergence on high-frequency information being much slower. This issue can result in key details being smoothed out or lost during the projection process when handling visual features rich in intricate details.

To overcome these limitations, the TaiChi framework introduces KAN \cite{liu2024kan} as an expressive alternative projector. The design of KAN is inspired by the Kolmogorov-Arnold representation theorem, which states that any multivariable continuous function $f(\cdot)$ defined on a bounded domain can be represented as a finite sum of composite univariate continuous functions. 

\begin{figure}[htpb]
	\centering
	\includegraphics[width=5cm]{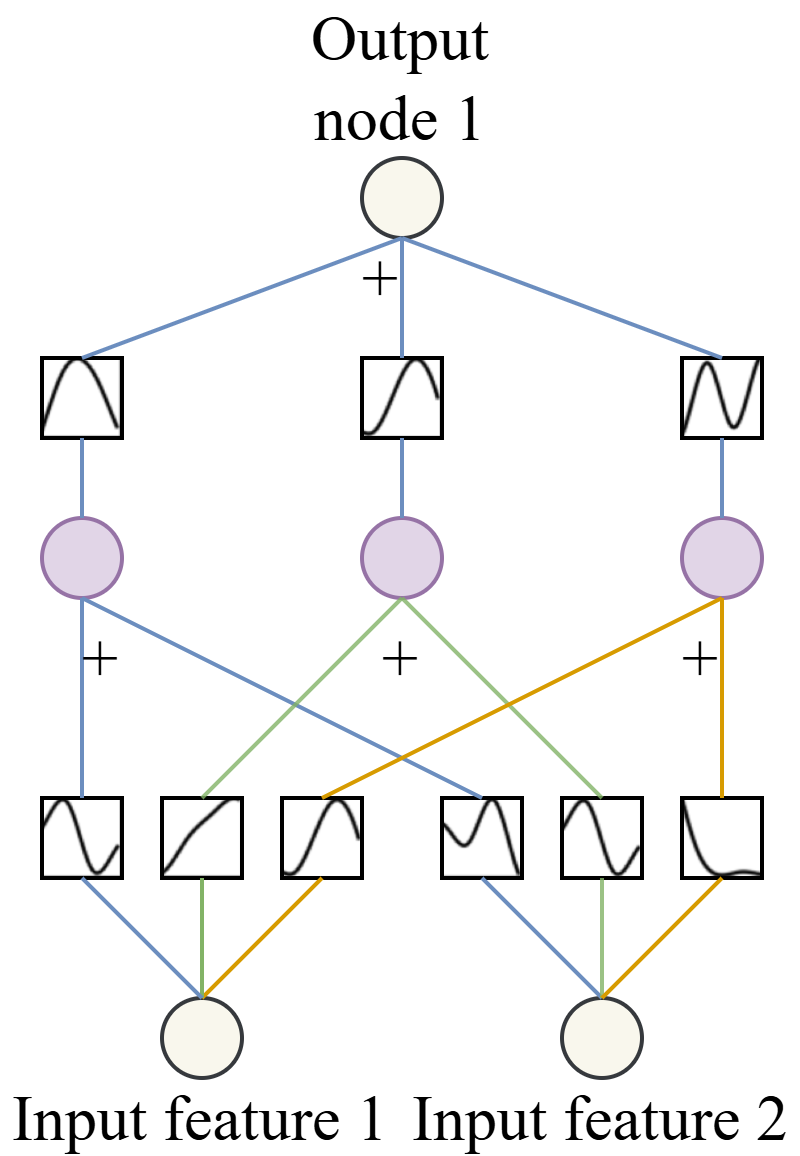}
	\caption{The structure of KAN.}
	\label{fig:kan}
\end{figure}

A Kolmogorov-Arnold layer with $d_{in}$ dimensional input and $d_{out}$ dimensional output is represented as:
\begin{equation}
\label{eq:kan}
\begin{split}
f(\textbf{x}) = \Phi \circ \textbf{x} = \left[ \sum_{i=1}^{d_{in}} \phi_{i,1}(x_i) \cdots \sum_{i=1}^{d_{in}} \phi_{i,d_{out}}(x_i) \right], \\  \Phi = \begin{bmatrix} \phi_{1,1}(\cdot) & \cdots & \phi_{1,d_{in}}(\cdot) \\ \vdots & \ddots & \vdots \\ \phi_{d_{out},1}(\cdot) & \cdots & \phi_{d_{out},d_{in}}(\cdot) \end{bmatrix},
\end{split}
\end{equation}
where $\textbf{x}=\left [ x_1,x_2,...x_i,... \right ] $ is the input data, and $\phi(\cdot)$ is a simple function. In practice, the activation functions SiLU and B-spline \cite{193220} are used to combine the linear parameters of the functions:
\begin{equation}
\phi(x) = w_b \text{SiLU}(x) + w_s \text{Spline}(x),
\end{equation}
\begin{equation}
\text{SiLU}(x) = \frac{x}{1 + e^{-x}}, \quad \text{Spline}(x) = \sum_i c_i B_i(x),
\end{equation}
where $\text{Spline}(\cdot)$ is the B-spline function, $w_b$ and $w_s$ represent the parameters of the SiLU activation and the B-spline basis function, and $c_i$ are the weights of the B-spline functions, $B_i(x)$ being the B-spline basis functions. The structure of KAN is shown in Fig. \ref{fig:kan}.

\subsection{Instructional Diffusion Refinement}
Although directly using the textual output of an LLM as the prompt input for a diffusion model is an intuitive approach to achieving end-to-end multimodal output in VLMs, this strategy faces notable performance bottlenecks due to the “semantic gap” between the LLM’s text generation logic and the generation requirements of diffusion models. The text encoder in diffusion models is typically trained using CLIP-based image–text contrastive learning, which effectively captures high-level semantic concepts but falls short in providing the fine-grained generation instructions—such as object counts, spatial arrangements, and complex attribute combinations—required for precise image synthesis.
To address this, the TaiChi framework proposes an LLM-driven prompt optimization strategy, termed IDR. By leveraging the LLM’s capabilities in language comprehension, reasoning, and rewriting, IDR refines and expands the initial textual instructions internally, without retraining the diffusion model, thereby enhancing generation quality and performance.

The specific implementation of IDR is a two-stage internal process. First, upon receiving the user's instruction for a non-text modality output (e.g., image, audio, video), the LLM generates a preliminary text output. Then, instead of directly passing the text output to the diffusion model, the system invokes a preset, structured instruction template, initiating an internal ``prompt self-optimization” task for the LLM. This template serves as a detailed set of meta-instructions, guiding the LLM to reconstruct the text according to a series of optimization principles. These principles include instructing the LLM to clearly articulate the core objectives, break down complex tasks into simpler descriptions, and use specific rather than abstract instructions. The template encourages the LLM to enrich key details in the prompt while maintaining the original intent and tone.

\section{TaiChi-Driven Token Communications}

\begin{figure*}[htpb]
	\centering
	\includegraphics[width=18cm]{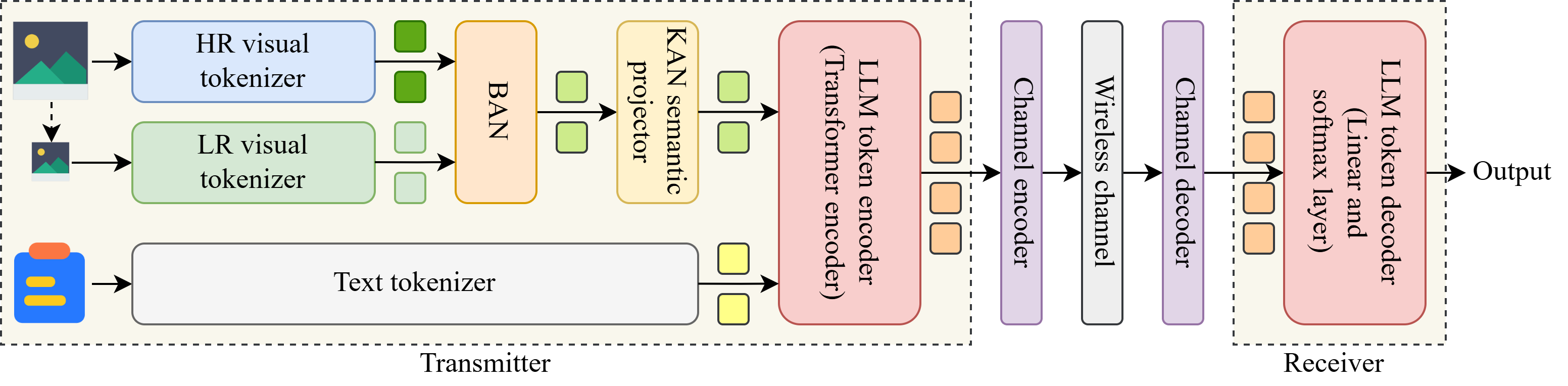}
	\caption{The structure of the TaiChi-driven token communication system.}
	\label{fig:sys}
\end{figure*}
Token communication is an AI-native communication paradigm that uses learnable semantic tokens as the fundamental transmission units\cite{11175596}. The core idea is to transform multimodal signals into structurally consistent token sequences through a unified tokenizer, thereby aligning communication representations naturally with the reasoning processes of VLMs. In this framework, the system no longer relies on bit-level encoding; instead, compression, transmission, and reconstruction are all performed at the token level, enabling a tight integration of communication functionalities with the intelligent reasoning capabilities of VLMs. A comparison between Token communication and conventional semantic communication is summarized in Table~\ref{tab:algo_performance}.

Leveraging the strengths of the TaiChi framework in multimodal processing, this study develops a multimodal and multitask token communication system and systematically clarifies the functional roles of its key components within the token communication pipeline, including how the dual visual tokenizers, the BAN, and the KAN projectors respectively handle visual feature modeling and cross-modal projection, while collaboratively operating with the LLM to form an efficient end-to-end token communication framework. 

\begin{table*}[t]
	\centering
	\caption{Comparison Between TokCom and Semantic Communication}
	\label{tab:algo_performance}
	\renewcommand{\arraystretch}{1.2}
	\small
	\begin{tabular}{p{4.1cm} p{6.5cm} p{6.4cm}}
		\toprule
		\textbf{Comparison Dimension} 
		& \textbf{Token Communication} 
		& \textbf{Semantic Communication} \\
		\midrule
		Information Unit 
		& Discrete tokens serving as semantic symbols, generated via LLM/VLM-based semantic compression 
		& Low-dimensional semantic features or embeddings extracted by task-specific neural networks \\
		
		Encoding and Decoding 
		& Tokenization and detokenization with generative semantic reconstruction 
		& Semantic encoding and decoding through end-to-end trained models \\
		
		Relation to LLMs/VLMs 
		& Strong dependence on LLMs/VLMs for semantic understanding, reasoning, and generation 
		& Large models are optional; typically relies on task-driven semantic compression networks \\
		
		Cross-Modal Support 
		& Native cross-modal support (text, image, audio) via unified token representation 
		& Modality-specific encoders; lacks a unified cross-modal semantic representation \\
		
		Transmission Robustness 
		& High robustness enabled by generative contextual completion and token-level recovery 
		& Robustness achieved mainly via end-to-end training; no explicit generative recovery mechanism \\
		
		Task Generalization 
		& Strong generalization benefiting from LLM/VLM knowledge and multi-task reasoning capability 
		& Limited generalization due to task-specific training and weak transferability \\
		
		System Design Complexity 
		& High complexity involving tokenizers, LLMs/VLMs, and token-level decoding pipelines 
		& Moderate complexity with a single semantic encoder–decoder architecture \\
		\bottomrule
	\end{tabular}
\end{table*}

\subsection{System Model}
The proposed token communication system consists of three main components: a transmitter, a receiver, and a physical channel, as shown in Fig. \ref{fig:sys}.

\subsubsection{Transmitter}

The first stage of the transmitter focuses on extracting and aligning token information from the visual modality.
Given the input image $\mathbf{I} \in \mathbb{R}^{H \times W \times C}$, a downsampled image version $\mathbf{I}'$ is first generated to capture complementary global features. Two visual tokenizers, denoted as $V_l(\cdot)$ and $V_h(\cdot)$, are employed to extract multi-level visual token representations from $\mathbf{I}$ and $\mathbf{I}'$, respectively. The outputs of the two visual tokenizers are then fused and enhanced by BAN $B_{\varepsilon}(\cdot)$ with parameter set $\varepsilon$, which effectively models the fine-grained interactions between different visual tokens. Subsequently, a KAN projector $P_{\zeta}(\cdot)$ with parameter set $\zeta$ is introduced to map the fused visual token representation into the textual semantic space, enabling alignment with language-based representations. The resulting visual token sequence $\mathbf{Z}_v$ can be expressed as
\begin{equation}
	\mathbf{Z}_v = P_{\zeta}\left( B_{\varepsilon}\left( V_l(\mathbf{I}'), V_h(\mathbf{I}) \right) \right).
\end{equation}

In parallel with visual processing, the textual query $t$ is transformed into a vector representation using a text tokenizer $E_{\theta}(\cdot)$ with parameter set $\theta$. This text tokenizer converts the input text into a sequence of language tokens residing in the same semantic space as the projected visual tokens. The multimodal token encoder $S_{\alpha}(\cdot)$ with parameter set $\alpha$ then integrates the visual  tokens $\mathbf{Z}_v$ and the textual  tokens $E_{\theta}(t)$, producing a unified multimodal token representation. This representation captures both the content of the image and the intent expressed by the textual query, and serves as the high-level token information to be transmitted. The high-level token sequence $\mathbf{Z}$ is given by
\begin{equation}
	\mathbf{Z} = S_{\alpha}\left( E_{\theta}(t), \mathbf{Z}_v \right).
\end{equation}

The final stage of the transmitter performs channel encoding to ensure robust transmission of the multimodal token representation over the physical channel. The channel encoder $C_{\beta}(\cdot)$ with parameter set $\beta$ maps the  token sequence $\mathbf{Z}$ into a sequence of channel symbols $\mathbf{Y}$, which are suitable for transmission under practical channel constraints. The encoded symbol sequence can be written as
\begin{equation}
	\mathbf{Y} = C_{\beta}(\mathbf{Z}).
\end{equation}
\subsubsection{Physical channel}
The transmitter sends the encoded symbols $\textbf{Y}$, which are transmitted through the physical channel to the receiver. The channel output sequence $\hat{\textbf{Y}} $ at the receiver can be expressed as:
\begin{equation}
\hat{\textbf{Y}}  = \textbf{H}\textbf{Y} + \textbf{n},
\end{equation}
where $\textbf{H}$ represents the channel gain, and $\textbf{n}$ is the Additive White Gaussian Noise (AWGN).

\subsubsection{Receiver}
Similar to the transmitter, the receiver consists of two components: a channel decoder and a token decoder. The channel decoder and token decoder are responsible for converting the received symbols into token information and generating the final textual output based on the token information, respectively. This process can be expressed as:
\begin{equation}
o = S_\delta^{-1} (C_\gamma^{-1} (\hat{\textbf{Y}})),
\end{equation}
where $C_\gamma^{-1}(\cdot)$ is the channel decoder with parameter set $\gamma$, and $S_\delta^{-1}(\cdot)$ is the token decoder with parameter set $\delta$.

To maintain the performance of the VLM, it is crucial to preserve the semantic consistency between $o$ and $\hat{t}$, where $\hat{t}$ represents the true answer corresponding to the question $t$. We use Cross-Entropy (CE) as the loss function:
\begin{equation}
\label{eq:sc_loss}
L_{\text{CE}}(o, \hat{t}) = - \sum_{l=1}^{L} q_{w_l} \log p_{w_i} + (1 - q_{w_l}) \log (1 - p_{w_i}),
\end{equation}
where $q_{w_l}$ represents the true probability of the $l$-th word $w_l$ in $\hat{t}$, and $p_{w_i}$ represents the predicted probability of the $l$-th word $w_i$ in $o$. CE is used to measure the difference between the two probability distributions. By minimizing the CE loss, the system can learn the meaning of the true answer $\hat{t}$ corresponding to the question $t$ in terms of its semantics, logic, and conceptual information. Therefore, the objective of the token communication system is to determine the parameters of the token encoder/decoder, channel encoder/decoder, BAN, and KAN projector, denoted as $\alpha^*, \beta^*, \gamma^*, \delta^*, \varepsilon^*, \zeta^*$, in order to minimize the expected distortion, as shown below:
\begin{equation}
(\alpha^*, \beta^*, \gamma^*, \delta^*, \varepsilon^*, \zeta^*) = \arg \min_{\alpha, \beta, \gamma, \delta, \varepsilon, \zeta} \mathbb{E}_{P(o, \hat{t})}[L_{CE}(o, \hat{t})],
\end{equation}
where $\alpha^*$ represents the optimal token encoder parameters, $\beta^*$ represents the optimal channel encoder parameters, $\gamma^*$ represents the optimal channel decoder parameters, $\delta^*$ represents the optimal token decoder parameters, $\varepsilon^*$ represents the optimal BAN parameters, and $\zeta^*$ represents the optimal projector parameters. $P(o, \hat{t})$ denotes the joint probability distribution of $o$ and $\hat{t}$.

\subsection{Visual Token Compression}

From the token communication perspective, BAN plays a central role in high-quality visual token compression. Rather than acting as a simple feature fusion module, BAN functions as a dynamic and task-driven token compressor and enhancer. By adaptively integrating critical fine-grained details into global conceptual representations according to the implicit requirements of the communication task, BAN effectively suppresses redundant visual information while emphasizing task-relevant components. As a result, BAN produces compact yet information-rich visual tokens, enabling high-fidelity and high-efficiency end-to-end token transmission.

\subsection{Cross-Modal Token Alignment}
Following visual token fusion, KAN serves as a high-fidelity projector to achieve accurate cross-modal alignment. By leveraging learnable activation functions, KAN adaptively constructs complex nonlinear mappings from visual features to text embeddings, significantly reducing information loss and distortion during modality transformation. This design ensures that the fine-grained details and complex spatial relationships preserved by BAN are faithfully transmitted to the LLM, providing more precise and comprehensive token inputs for subsequent understanding and reasoning, and ultimately enhancing the system’s accuracy and robustness in complex multimodal communication tasks.

\subsection{LLM Token Encoding}

In the token communication system, the LLM is responsible for the deep token understanding of multimodal inputs. To fit the transmitter-receiver architecture, the LLM is decomposed into a token encoder and a token decoder, forming the intelligent core of the transmission channel. At the transmitter, multimodal tokens are fed into multiple Transformer encoder layers, where multi-head self-attention mechanisms and MLP networks enable cross-modal feature fusion and modeling of complex intrinsic relationships between tokens. As a result, the token encoder generates a unified, high-level multimodal token representation that encapsulates the core intent of the transmitted information and serves as the input to the channel encoder.

\subsection{LLM Token Decoding}
At the receiver, the token decoder is responsible for reconstructing meaningful outputs from the received token representation. Specifically, the decoded token vector is mapped back to the vocabulary space through a linear projection followed by a Softmax function, producing a probability distribution over candidate tokens. Based on this distribution, the token decoder autoregressively generates human-readable text. This decoding process focuses on precise information generation, ensuring that the reconstructed output faithfully reflects the original intent conveyed by the transmitter. Together with LLM token encoding, this functional decomposition clearly defines the two key stages of token communication—deep understanding and encoding at the transmitter, and accurate decoding and generation at the receiver—forming an efficient end-to-end token communication loop.

From the perspective of token communication, the LLM endows the token encoder with high flexibility, with the task instruction serving as a high-level control signal for the LLM, guiding adaptive information extraction and encoding for different tasks. For instance, in VQA tasks, the encoder focuses on key objects and details, whereas in image captioning tasks, it emphasizes the overall scene and primary elements, thereby enhancing information relevance and optimizing communication resource utilization. Moreover, this strategy ensures end-to-end alignment of communication intent, enabling the token decoder’s output to more accurately reflect the transmitter’s original intent. Consequently, it elevates the TaiChi framework into a practical, flexible, and efficient multimodal multitask token communication system.

\begin{figure*}[htpb]
	\centering
	\includegraphics[width=17cm]{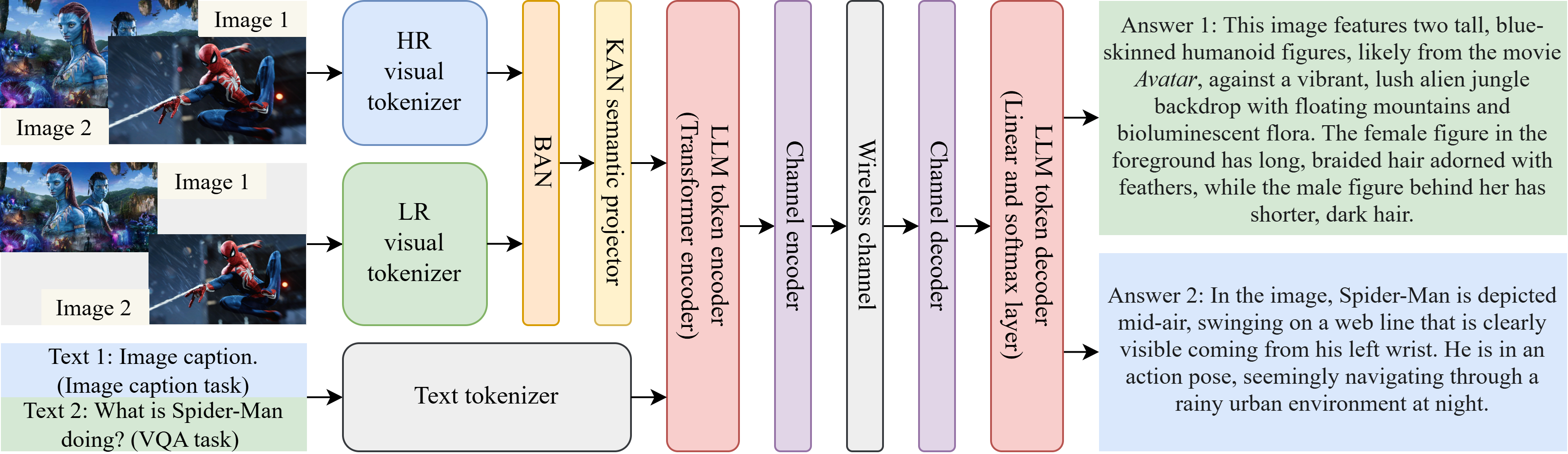}
	\caption{Different multimodal task examples for TaiChi.}
	\label{fig:task}
\end{figure*}

\section{Phase-wise Training}
The training of the entire system is divided into two stages: the first stage focuses on training the TaiChi, while the second stage conducts joint encoding training of the VLM and the channel module within the token communication framework; during the first stage, a vision–language multitask instruction fine-tuning strategy is introduced to enhance the system’s adaptability to diverse tasks, enabling it to interpret natural language instructions and flexibly perform different communication tasks, such as VQA and image captioning, thereby achieving a unified multitask token communication capability.

\subsection{VLM Training}
\begin{figure}[htpb]
	\centering
	\includegraphics[width=8.5cm]{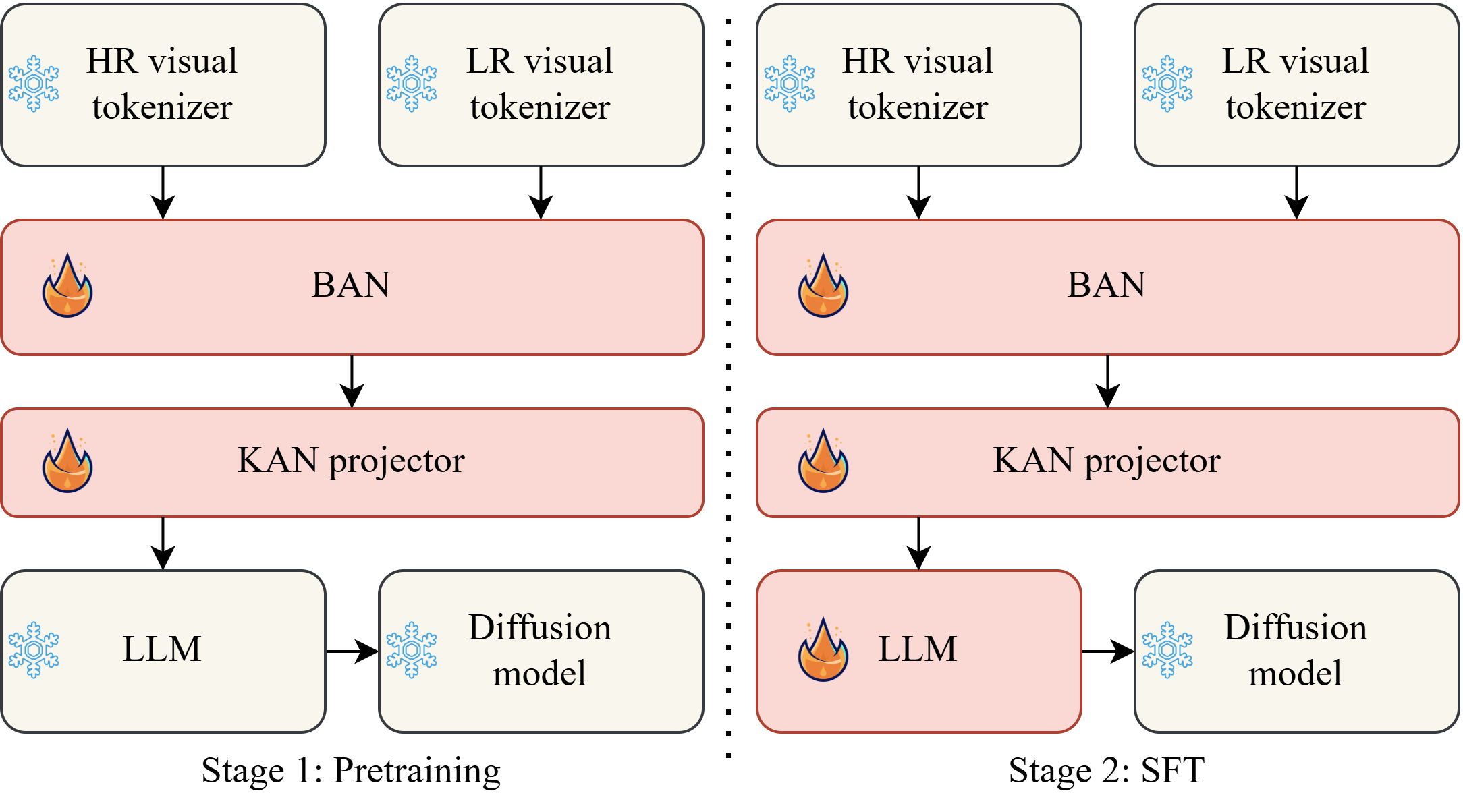}
	\caption{The two training stages of TaiChi.}
	\label{fig:vlm_train}
\end{figure}

The training of TaiChi is carried out in two stages, progressively introducing higher-quality data and gradually unfreezing more parts of the model to improve the efficiency and stability of the training process, as shown in Fig. \ref{fig:vlm_train}.

\subsubsection{Pretraining}
The primary objective of this stage is to calibrate the backbone network and initialize the training of newly introduced parameters, thereby providing stable feature representation capabilities and enhancing robustness to out-of-distribution data for subsequent tasks. During training, only the BAN and KAN projectors are updated, while all other components remain frozen. This process can be regarded as training a compatible visual tokenizer for the frozen LLM. The datasets used include LAION \cite{schuhmann2022laion} and ALLaVA \cite{chen2024allava}, which cover large-scale cross-modal corpora to ensure that the model acquires extensive visual–language knowledge. The loss function is the CE loss, computed solely on the response part.

\subsubsection{Supervised fine-tuning (SFT)}

To enhance the flexibility and adaptability of the TaiChi-driven token communication system across diverse tasks, we adopt a multimodal and multitask instruction fine-tuning strategy. The core idea is to unify different multimodal tasks under a structured instruction format, enabling positive knowledge transfer between tasks while mitigating catastrophic forgetting in multitask learning, thereby improving the system’s multifunctionality and generalization capability \cite{wei2021finetuned}.

In practice, we design natural language instruction templates for various communication tasks (e.g., image captioning, VQA), specifying the task objective, context, and expected output format. During training, multitask data are mixed with diverse instructions, compelling the model to not only master task-specific solutions but also acquire the ability to follow instructions, thus facilitating cross-task capability transfer and enhancing overall performance. As shown in Fig. 6, when a user initiates a task, their request is presented in a structured instruction format, for example:

For the image captioning task, the full dialogue with instructions may be as follows:

\textit{\textbf{Instruction}: ``Please carefully observe the provided image and identify the main elements, including the scene (e.g., indoor, outdoor, city, nature, etc.), key objects (e.g., animals, people, buildings, tools, etc.), and any actions or activities (e.g., running, eating, reading, etc.). Based on this information, generate a natural language description that covers the key details of the image. Ensure the description is fluent, grammatically correct, and avoids excessive speculation, only describing what can be clearly observed in the image."
	\textbf{Input Image}: Image 1.
	\textbf{Input}: ``Image caption."
	\textbf{Output}: ``This image features two tall, blue-skinned humanoid figures, likely from the movie Avatar, against a vibrant, lush alien jungle backdrop with floating mountains and bioluminescent flora. The female figure in the foreground has long, braided hair adorned with feathers, while the male figure behind her has shorter, dark hair."}

For the VQA task, the full dialogue with instructions may be as follows:

\textit{\textbf{Instruction}: ``Please observe the provided image and read the question carefully to understand its specific requirements. Based on the content of the image, answer the question, ensuring that the answer is based on clearly observable information. If the question involves specific details (e.g., quantity, color, location), provide precise answers. If the question cannot be answered based on the image, respond with: `Cannot obtain an answer from the image.' Finally, provide a concise and clear answer that fully satisfies the requirement of the question."
	\textbf{Input Image}: Image 2.
	\textbf{Input}: ``What is Spider-Man doing?"
	\textbf{Output}: ``In the image, Spider-Man is depicted mid-air, swinging on a web line that is clearly visible coming from his left wrist. He is in an action pose, seemingly navigating through a rainy urban environment at night."}

A portion of the datasets is sourced from Cauldron \cite{laurenccon2024matters}, formatted in a standardized question–answer structure, where multiple Q\&A pairs for a single image are organized into multi-turn dialogues. The CE loss is again employed, computed only on the response part, to ensure that optimization focuses on improving output quality.
\subsection{Joint VLM-channel coding}
\begin{figure}[htpb]
	\centering
	\includegraphics[width=5.5cm]{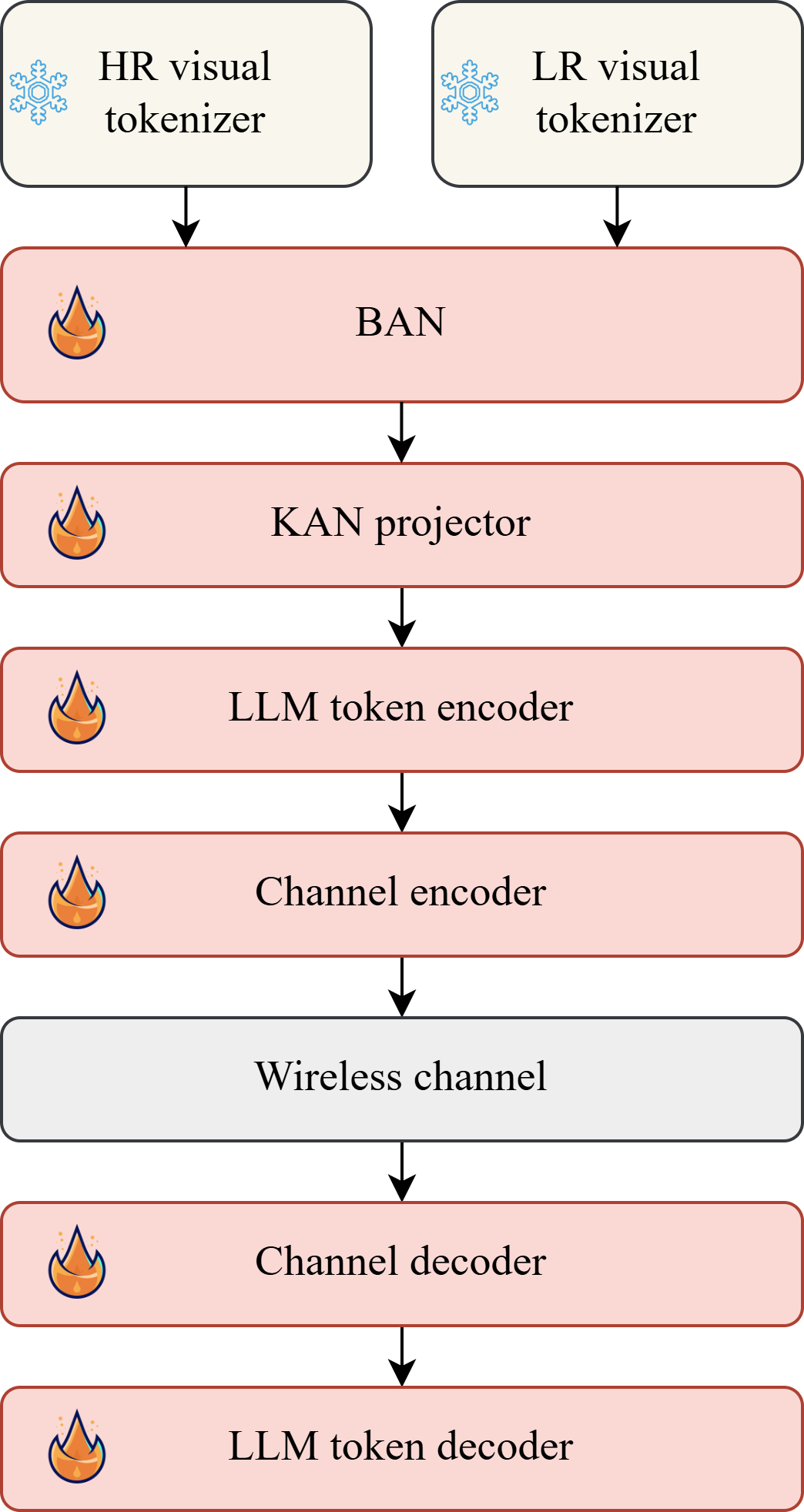}
	\caption{The joint coding process of TaiChi-driven token communication system.}
	\label{fig:jc}
\end{figure}

In the joint coding phase, we combine the optimized TaiChi model with the channel encoder-decoder for joint coding. The focus of this phase is to improve the overall performance of the token communication system. During the joint VLM-channel coding process, we allow the BAN, KAN, and channel encoder-decoder to undergo full updates to adapt to complex channel conditions. Meanwhile, we fine-tune the LLM using LoRA \cite{hu2022lora}. This strategy significantly reduces training costs while maintaining model performance, as shown in Fig. \ref{fig:jc}. 

\section{Experimental Results}
In this section, we evaluate the TaiChi model and the TaiChi-driven token communication system across multiple benchmarks, providing an analysis of each component along with qualitative results.

We trained two versions of TaiChi, using the LLMs Gemma-2B-instruct and Qwen-2.5-14B-instruct \cite{qwen2.5}, respectively. Meanwhile, the model used by the LR visual tokenizer is CLIP-ViT-large-patch14-336 \cite{radford2021learning}, while the model used by the HR visual tokenizer is CLIP-ConvNeXt\_large\_d\_320 \cite{liu2022convnet}.

Regarding the training setup, we employed the AdamW optimizer and a cosine learning rate schedule, with one round of optimization for all models. The initial learning rates for pretraining and SFT were set to 1e-3 and 2e-5, respectively, to ensure training stability. The global batch sizes for pre-training and SFT were set to 256 and 128, respectively. Training and testing were conducted on a server equipped with an Intel Xeon CPU (2.6 GHz, 1007 GB RAM) and 8 NVIDIA A800 GPUs (80 GB SGRAM). The training framework used was PyTorch, and the distributed training strategy employed was DeepSpeed ZeRO-2.

\subsection{Zero-Shot Evaluation across Multimodal Benchmarks}
To comprehensively evaluate the core performance of the TaiChi framework under zero-shot conditions, we conducted comparative experiments across multiple widely recognized multimodal benchmarks. This evaluation aims to verify the superiority of TaiChi’s model design and its generalization capability in handling diverse tasks.

For an objective assessment, we compared TaiChi against several state-of-the-art VLMs, including the lightweight MobileVLM \cite{chu2023mobilevlm}, as well as widely used models such as InstructBLIP \cite{dai2023instructblip}, BLIP-2 \cite{li2023blip2}, and LLaVA-1.5 \cite{liu2024improved}. The performance evaluation spans multiple authoritative multimodal benchmarks, including VQA$^T$ \cite{singh2019towards}, MMB \cite{liu2024mmbench}, and MME \cite{fu2024mme},
which collectively assess the model’s capabilities in VQA, multimodal understanding, and reasoning.

As shown in Table \ref{tab:eval}, the experimental results clearly demonstrate the superior performance of the TaiChi framework. When equipped with Gemma-2B as its LLM, TaiChi not only outperforms the efficient MobileVLM but also surpasses InstructBLIP models powered by significantly larger LLMs, such as Vicuna-7B and Vicuna-13B. Furthermore, when integrated with the more powerful Qwen2.5-14B LLM, TaiChi exhibits strong scalability. Under equivalent LLM configurations, TaiChi consistently achieves significantly higher scores than LLaVA-1.5 across all benchmarks. The outstanding performance of TaiChi is primarily attributed to its innovative core architectural design, which effectively overcomes the limitations of existing VLMs in visual information processing and modality alignment.

\begin{table}[h]
	\centering
	\caption{Comparison on zero-shot benchmarks.}
	\label{tab:eval}
	\renewcommand{\arraystretch}{1.5}
	\begin{tabular}{|c|c|c|c|c|c|}
		\hline
		Method & LLM & Res. & VQA$^T$ & MMB & MME \\
		\hline
		MobileVLM & MLLaMA 2.7B & 336 & 47.5 & 59.6 & 1289 \\
		\hline
		BLIP-2 & Vicuna-13B & 224 & 42.5 & - & 1293.8 \\
		\hline
		InstructBLIP & Vicuna-7B & 224 & 50.1 & 36.0 & - \\
		\hline
		InstructBLIP & Vicuna-13B & 224 & 50.7 & - & 1212.8 \\
		\hline
		LLaVA-1.5 & Vicuna-13B & 336 & 61.3 & 69.2 & 1531/295 \\
		\hline
		TaiChi & Gemma-2B & 336 & 59.2 & 61.9 & 1323/240 \\
		\hline
		TaiChi & Qwen2.5-14B & 336 & \textbf{65.1} & \textbf{73.5} & \textbf{1565/339} \\
		\hline
	\end{tabular}
	\raggedright
	\textit{Note: Res. refers to the resolution of the input images.}
\end{table}

\subsection{Ablation Study on BAN and KAN}
To validate the effectiveness of the two core components, the BAN and the KAN projector, we conducted a series of ablation experiments. By replacing these modules with more basic alternatives, we assessed their individual contributions to the overall performance of the model. Specifically, we constructed two variants of TaiChi: (1) TaiChi (w/o KAN), in which the KAN projector is replaced with a traditional MLP projector; and (2) TaiChi (w/o BAN), where the BAN is removed. All variants were evaluated under the same Gemma-2B configuration as the full version of TaiChi. The evaluation datasets include VQA$^T$ \cite{singh2019towards}, MMB \cite{liu2024mmbench}, and MM-Vet \cite{yu2023mm}.

As shown in Table \ref{tab:abs}, removing any core component results in a significant performance drop across various benchmarks. Among the variants, TaiChi (w/o KAN) performed the worst, providing strong evidence of the key role of the KAN projector in achieving high-fidelity cross-modal alignment. Similarly, TaiChi (w/o BAN)'s performance was inferior to that of the full TaiChi model, highlighting the importance of BAN in efficiently fusing multi-scale visual features. In conclusion, these ablation studies clearly demonstrate that each component we proposed is crucial to the success of the TaiChi framework, and their collaboration collectively enhances the model's overall performance.

\begin{table}[t]
	\centering
	\caption{Ablation study of TaiChi's key components.}
	\label{tab:abs}
	\renewcommand{\arraystretch}{1.3}
	\small
	\begin{tabular*}{\columnwidth}{@{\extracolsep{\fill}}|l|c|c|c|}
		\hline
		Method & VQA$^T$ & MMB & MM-Vet \\
		\hline
		TaiChi & \textbf{59.2} & \textbf{65.1} & \textbf{33.3} \\
		\hline
		TaiChi (w/o KAN) & 55.0 & 60.7 & 24.5 \\
		\hline
		TaiChi (w/o BAN) & 59.0 & 60.9 & 28.5 \\
		\hline
	\end{tabular*}
\end{table}

\subsection{Token Communication Performance Evaluation}

\subsubsection{Multimodal and multitask evaluation}
\begin{figure}[h]
    \centering
    \begin{subfigure}[b]{0.45\textwidth}
        \includegraphics[width=\textwidth]{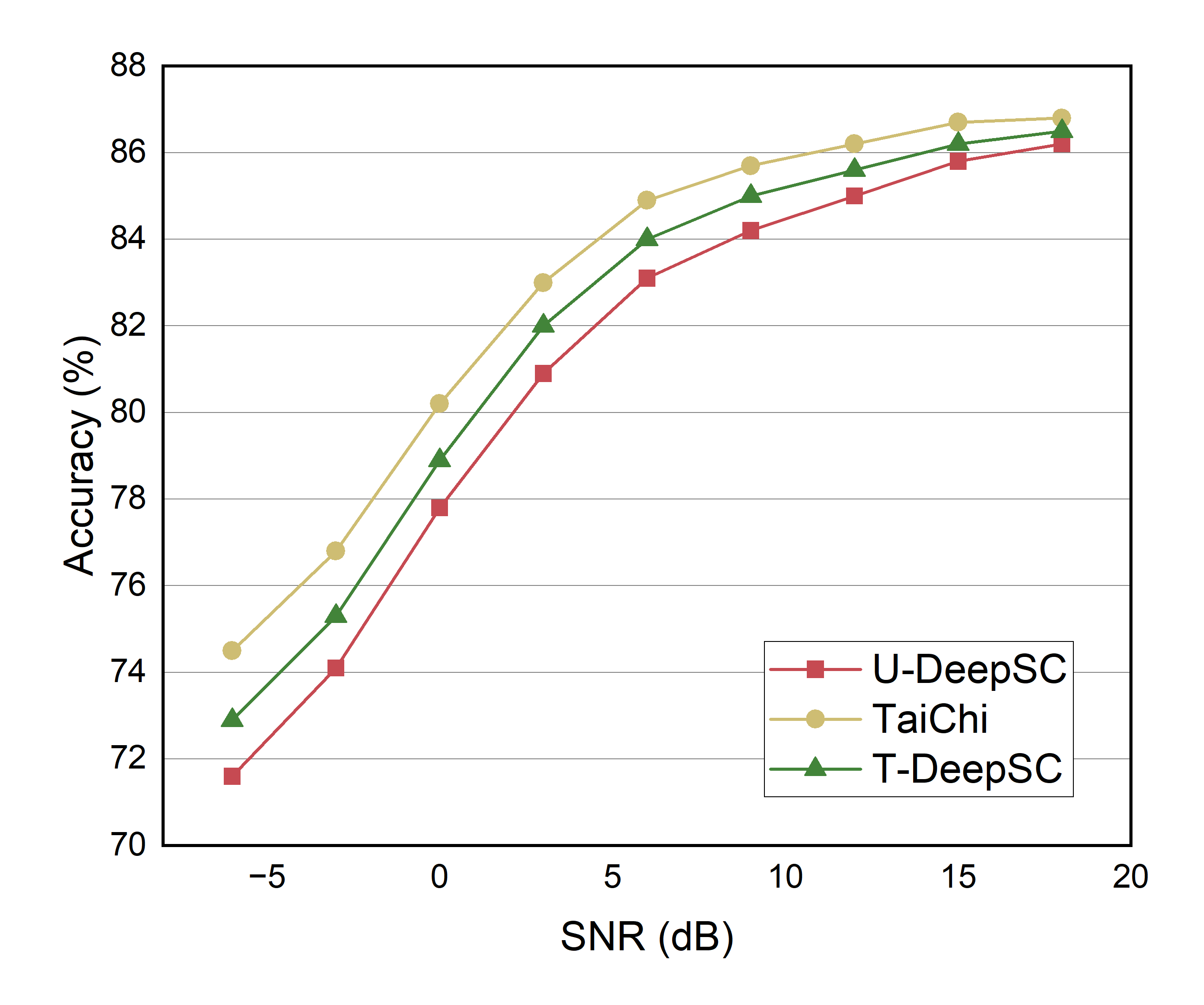} 
        \caption{AWGN}
        \label{fig:91}
    \end{subfigure}
    \hfill 
    \begin{subfigure}[b]{0.45\textwidth}
        \includegraphics[width=\textwidth]{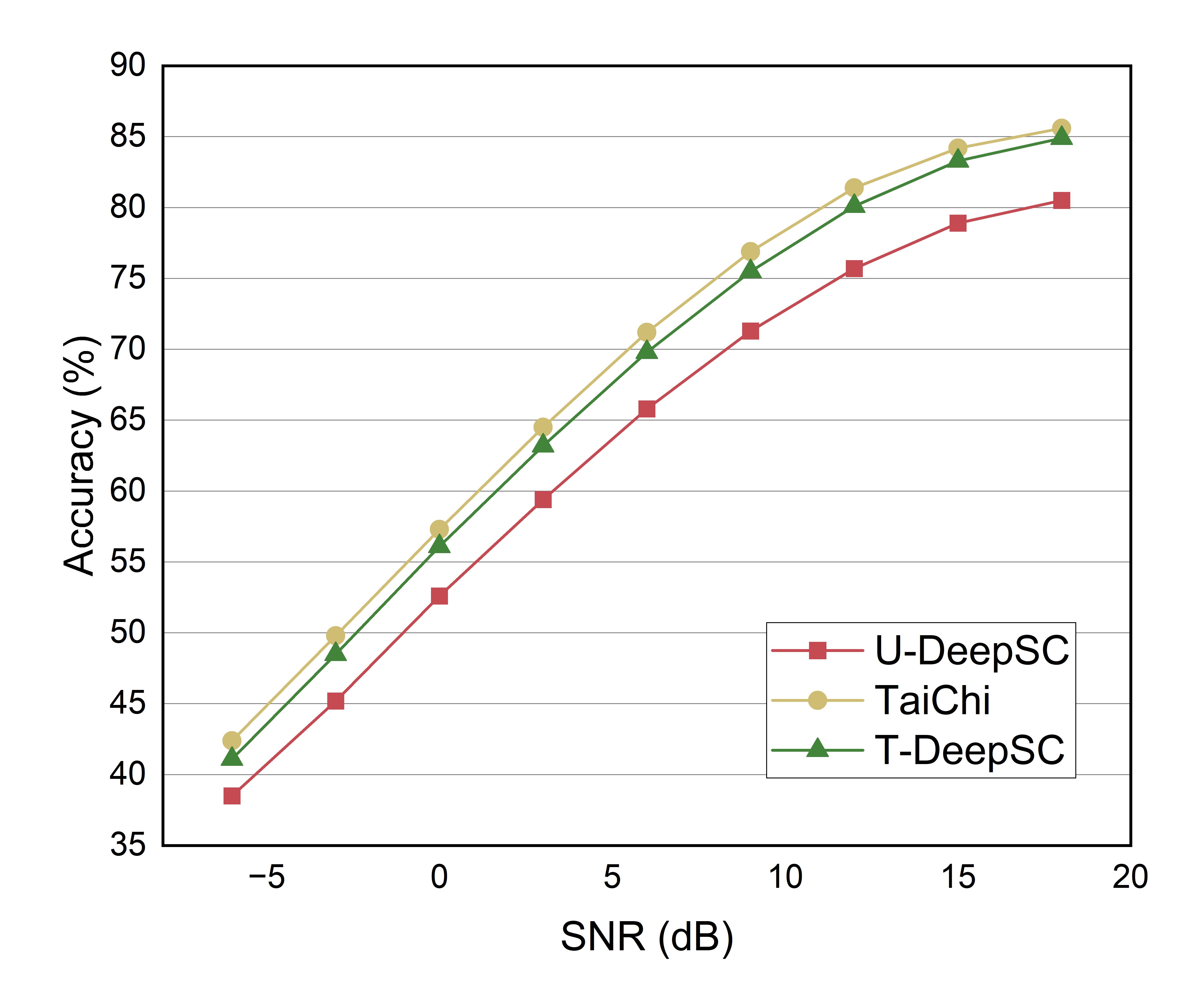} 
        \caption{Rayleigh fading}
        \label{fig:92}
    \end{subfigure}
    \caption{Performance of different tasks versus SNR: (a) VQAv2 dataset for VQA task; (b) SST2 dataset for text classification.}
    \label{fig:9}
\end{figure}

\begin{figure}[h]
    \centering
    \begin{subfigure}[b]{0.45\textwidth}
        \includegraphics[width=\textwidth]{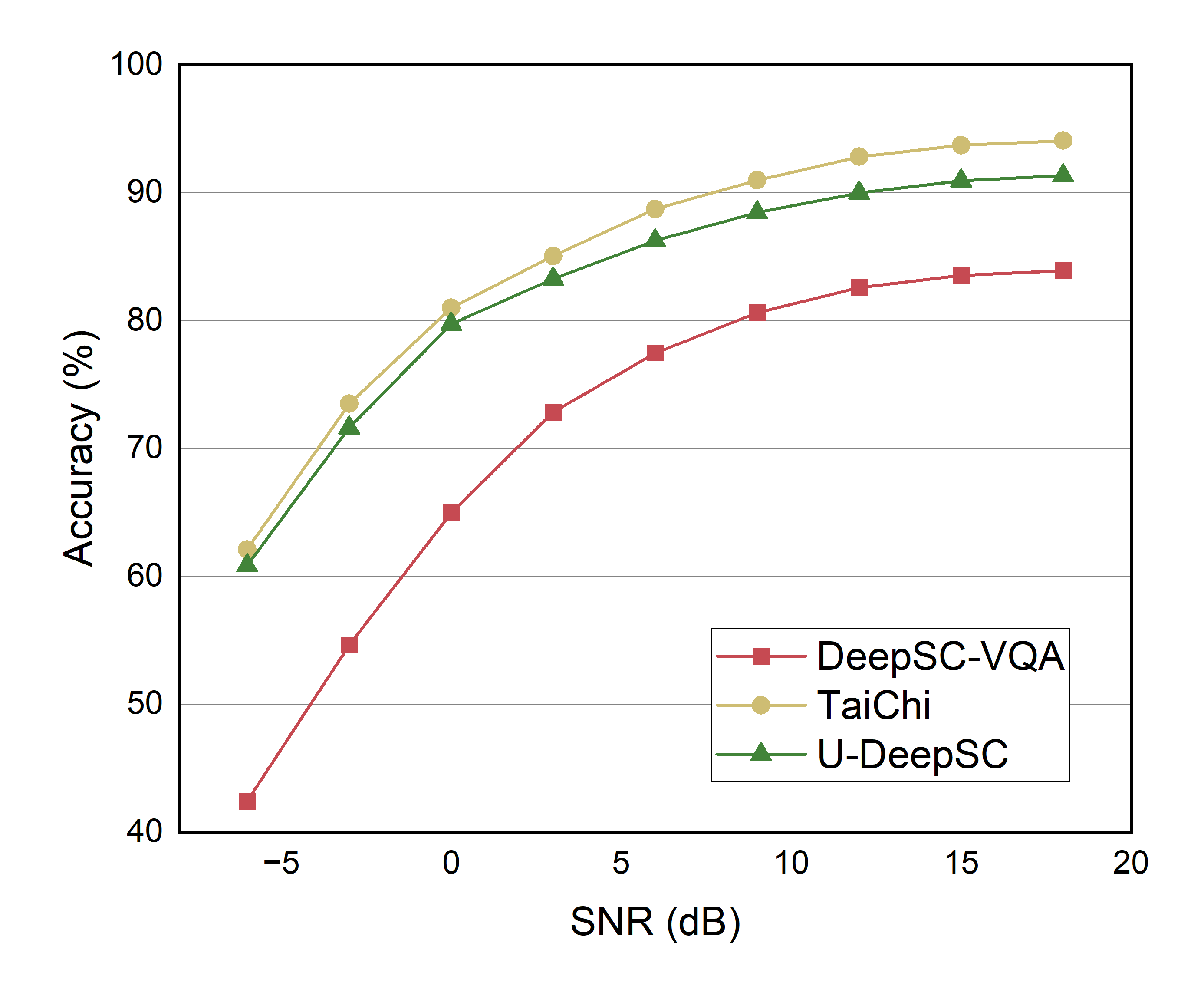} 
        \caption{AWGN}
        \label{fig:101}
    \end{subfigure}
    \hfill 
    \begin{subfigure}[b]{0.45\textwidth}
        \includegraphics[width=\textwidth]{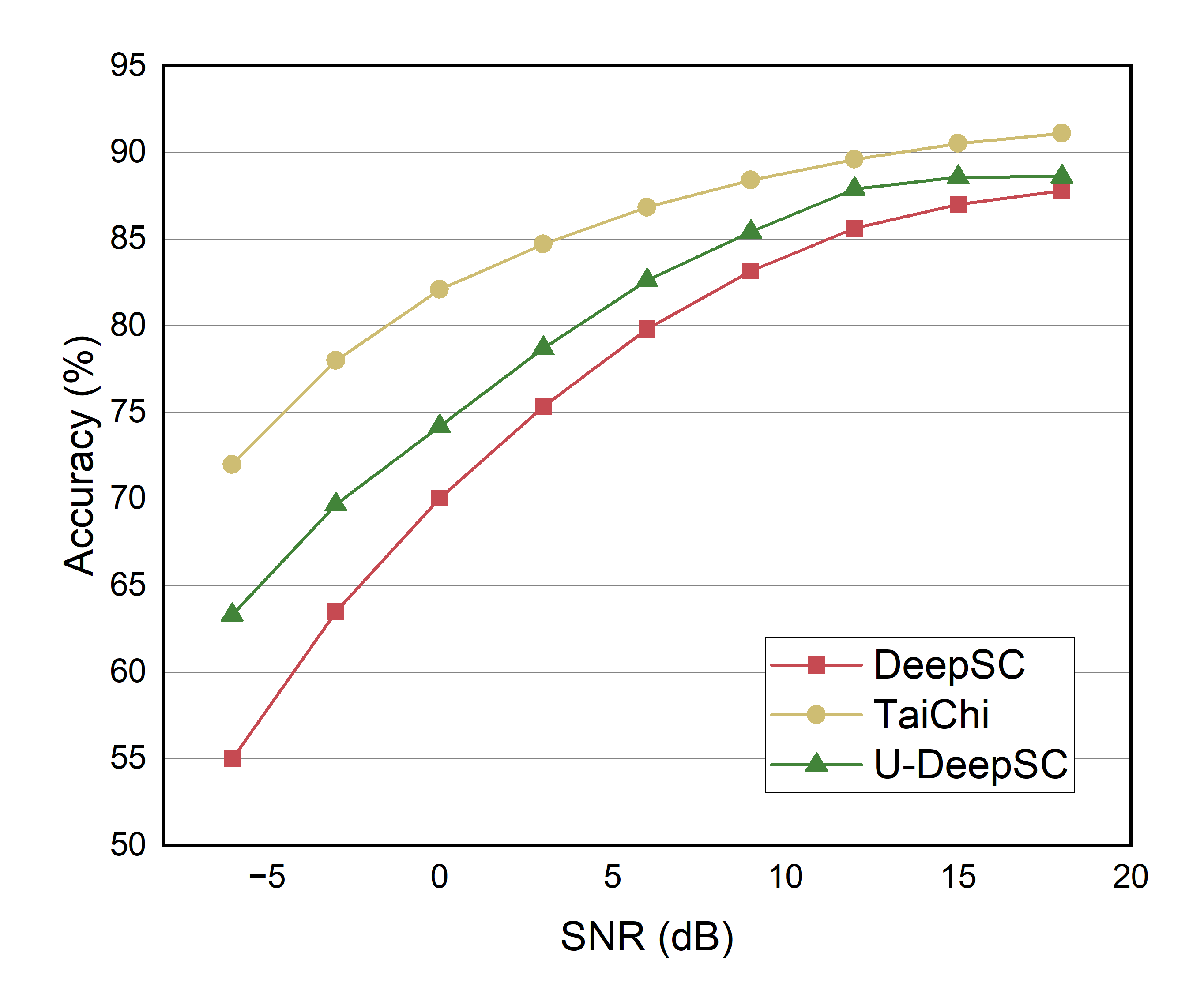} 
        \caption{Rayleigh fading}
        \label{fig:102}
    \end{subfigure}
    \caption{Performance of different tasks versus SNR: (a) CLEVR dataset for VQA task; (b) IMDB dataset for text classification.}
    \label{fig:10}
\end{figure}

To evaluate the multimodal performance of TaiChi, we compare it with U-DeepSC \cite{10431795}, T-DeepSC \cite{10431795}, and DeepSC-VQA \cite{9830752} on the VQA task. The datasets used for the VQA task are VQAv2 and CLEVR \cite{johnson2017clevr}. To assess TaiChi’s multitask performance, we conduct comparisons with U-DeepSC, T-DeepSC, DeepSC-VQA, and DeepSC \cite{9398576} across a range of downstream tasks, including VQA and text classification. The datasets used for the text classification task are SST-2 \cite{socher-etal-2013-recursive} and IMDB \cite{maas-EtAl:2011:ACL-HLT2011}. Accuracy is adopted as the evaluation metric for both experiments. The performance comparisons under the AWGN and Rayleigh channels for both multimodal and multitask scenarios are presented in Figs. \ref{fig:9} and \ref{fig:10}.

It can be observed that TaiChi consistently outperforms other models in both multimodal tasks such as VQA, and unimodal tasks such as text classification. Compared to token communication systems built on smaller models, TaiChi benefits from the powerful multimodal alignment and multitask learning capabilities of VLMs, demonstrating superior generalization performance in both multimodal and multitask settings.

\subsubsection{Performance evaluation of VLMs}
\begin{figure}[h]
    \centering
    \begin{subfigure}[b]{0.45\textwidth}
        \includegraphics[width=\textwidth]{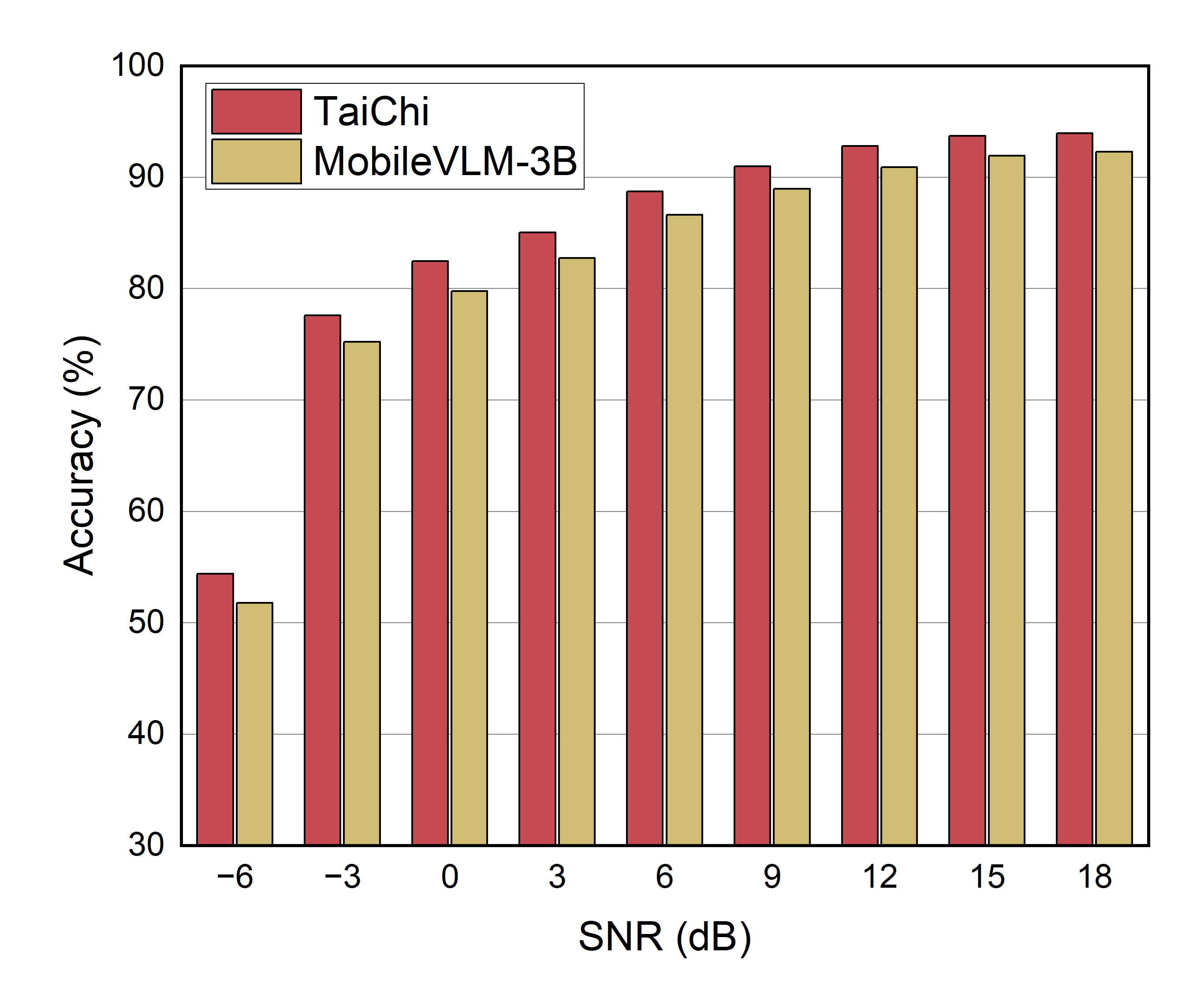} 
        \caption{AWGN}
        \label{fig:111}
    \end{subfigure}
    \hfill 
    \begin{subfigure}[b]{0.45\textwidth}
        \includegraphics[width=\textwidth]{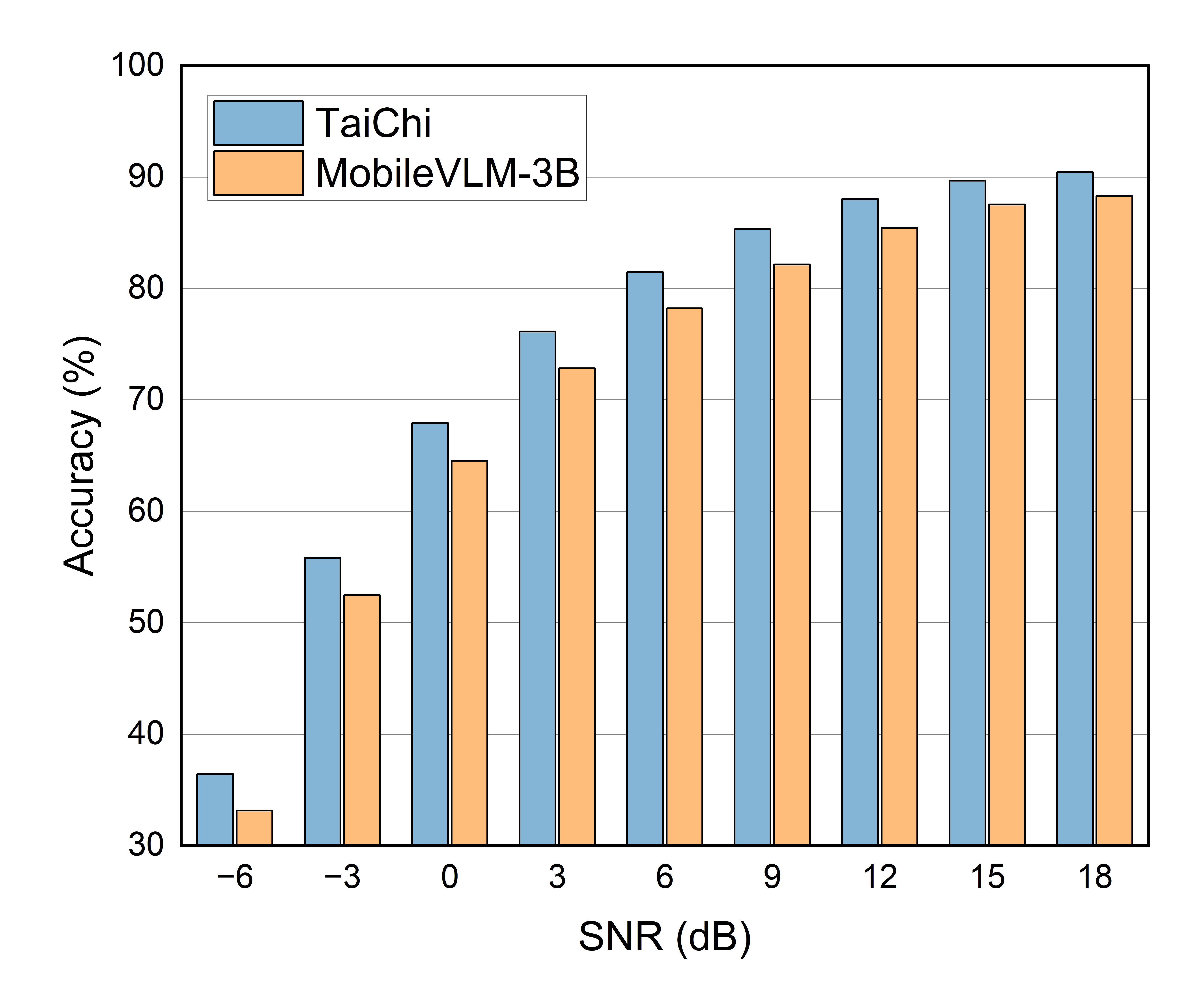} 
        \caption{Rayleigh fading}
        \label{fig:112}
    \end{subfigure}
    \caption{Performance of token communication systems with TaiChi and other VLMs.}
    \label{fig:11}
\end{figure}

To further validate the architectural advantages of the proposed TaiChi framework, we constructed a token communication system based on another VLM, MobileVLM-3B, for direct performance comparison. Both systems were evaluated under identical conditions on the CLEVR VQA task. To comprehensively assess performance robustness, the comparison was conducted under two standard channel models: AWGN channel and the more challenging Rayleigh fading channel. Fig. \ref{fig:11} presents the accuracy of both systems under varying SNR conditions.

It can be observed that, under both AWGN and Rayleigh fading channels, the TaiChi-driven token communication system consistently achieves significantly higher accuracy across all SNR levels compared to the system based on MobileVLM-3B. This result clearly demonstrates that the core design of the TaiChi framework, including the dual-visual tokenizers, BAN, and KAN projector, enables more effective extraction, fusion, and alignment of multimodal token information. As a result, TaiChi achieves stronger robustness and higher fidelity in end-to-end communication tasks, validating its superior performance as a token communication engine.

\subsubsection{Performance evaluation of joint coding }
To further verify the effectiveness of our proposed joint VLM-channel coding scheme, we conducted a comparative experiment aimed at assessing its contribution to the overall system performance. In this experiment, we evaluated the TaiChi-driven token communication system under two configurations: 1) a full system employing joint coding, in which the VLM and channel encoder-decoders are optimized collaboratively during training; and 2) a system without joint coding, where the token encoder and channel encoder are treated as separate, sequentially optimized modules. The accuracy of both configurations was tested on the CLEVR VQA task under both AWGN and Rayleigh fading channel conditions. The results are shown in Fig. 12.

It can be seen that, across all SNR conditions and under both the AWGN and the more challenging Rayleigh fading channels, the TaiChi system with joint coding consistently outperforms the system without joint coding. This result strongly supports the claim that jointly optimizing the VLM and channel encoding-decoding processes enables the system to learn token representations that are more robust to channel noise and fading. Such end-to-end joint optimization allows token information to be encoded with awareness of the channel's statistical characteristics, thereby enabling more accurate reconstruction at the receiver end. This significantly enhances the system's resilience to interference and improves the overall fidelity of token communication.

\section{Conclusion}
This study presented a high-performance VLM framework, TaiChi, designed to provide more intelligent and efficient support for token communication. The model has a dual-visual tokenizer structure, which collaboratively extracts both global conceptual information and fine-grained visual details from images, thereby enabling flexible visual representations tailored to communication tasks. To integrate the multi-scale features extracted by the dual-visual tokenizers, a bilateral attention mechanism was designed, which effectively reduces the number of tokens and enhances token representation. Additionally, a KAN-based projector was employed to achieve more precise nonlinear alignment between the visual and language modalities. Furthermore, TaiChi was integrated into a multimodal and multitask token communication system, and through a joint VLM and channels coding for optimization, it significantly improves transmission robustness under complex channel conditions. Experimental results demonstrated that TaiChi not only exhibits powerful expressive capabilities in multimodal modeling, but also shows superior effectiveness and interference resistance in token communication systems.

\begin{figure}[h]
	\centering
	\begin{subfigure}[b]{0.45\textwidth}
		\includegraphics[width=\textwidth]{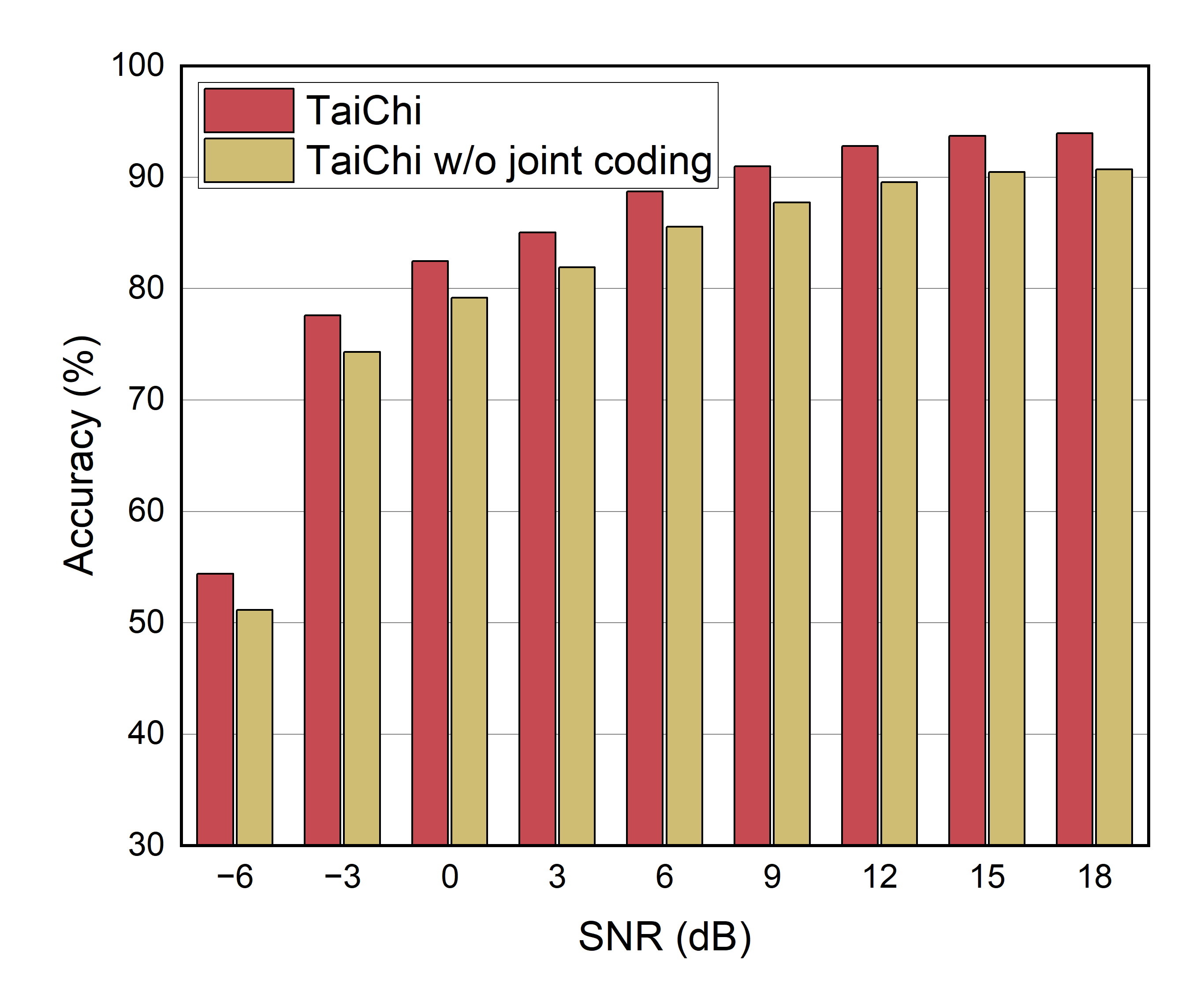} 
		\caption{AWGN}
		\label{fig:121}
	\end{subfigure}
	\hfill 
	\begin{subfigure}[b]{0.45\textwidth}
		\includegraphics[width=\textwidth]{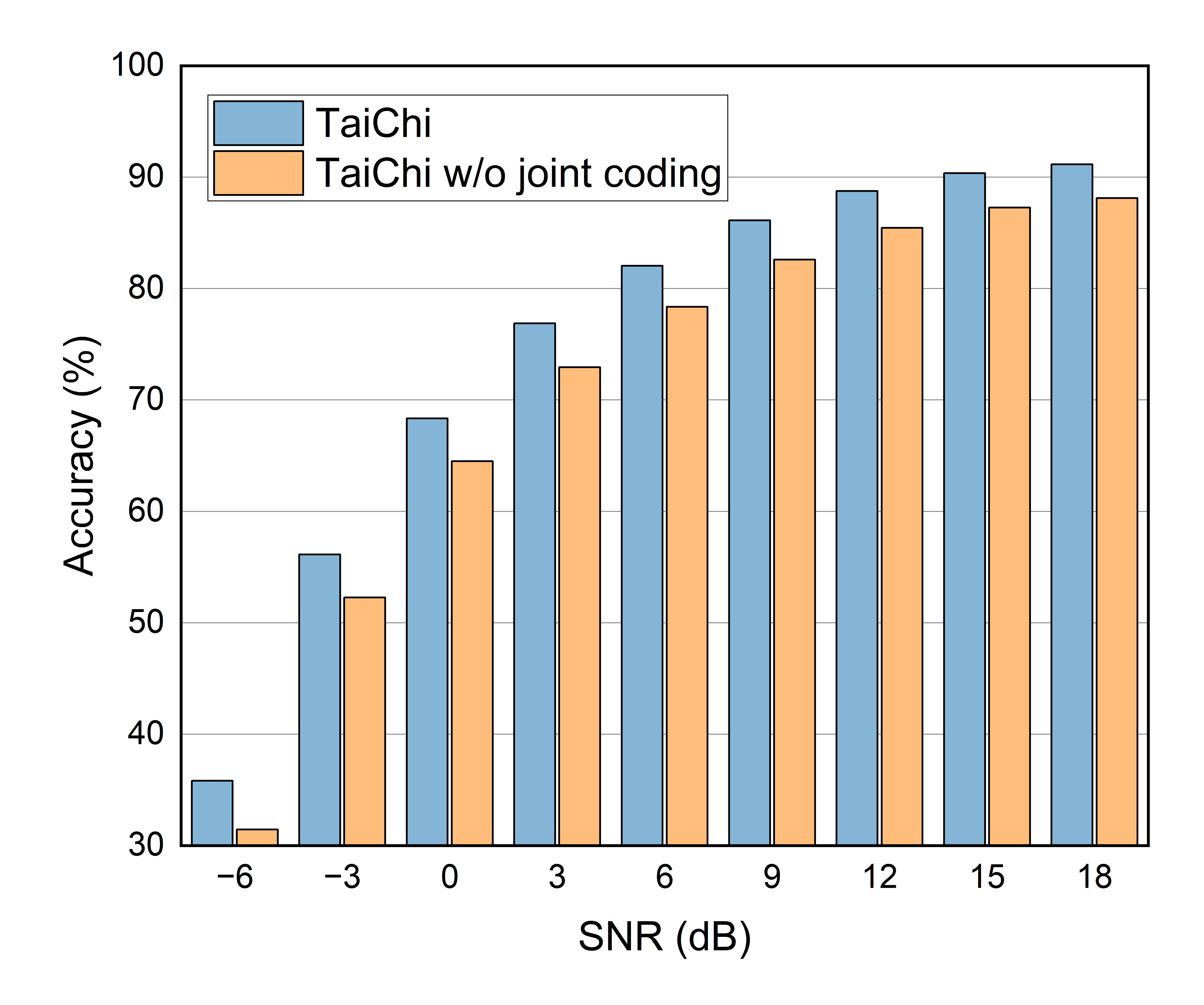} 
		\caption{Rayleigh fading}
		\label{fig:122}
	\end{subfigure}
	\caption{Performance of joint VLM-channel coding in token communication systems.}
	\label{fig:12}
\end{figure}

\bibliographystyle{IEEEtran}
\bibliography{bare_jrnl_bobo}

\end{document}